\documentclass[journal]{IEEEtran}

\usepackage{times}
\usepackage{epsfig}
\usepackage{graphicx}
\usepackage{amsmath}
\usepackage{amssymb}
\usepackage{authblk}
\usepackage{subfigure}
\usepackage{multirow}
\usepackage[percent]{overpic}
\usepackage{dblfloatfix}
\usepackage{xcolor}
\usepackage{rotating}
\usepackage{makecell}
\usepackage[export]{adjustbox}
\usepackage{hyperref}
\ifCLASSINFOpdf
\else
\fi
\begin{document}
%
\title{Adaptive Illumination based Depth Sensing using Deep Superpixel and Soft Sampling Approximation}
%
%
%

\author{Qiqin~Dai,	
        Fengqiang Li,
        Oliver Cossairt,
        and~Aggelos~K.~Katsaggelos,~\IEEEmembership{Fellow,~IEEE}
\thanks{Q. Dai is with Geomagical labs, Mountain View, CA, 94041, USA. F. Li is with Apple Inc, Cupertino, CA, 95014 . O. Cossairt and A. K. Katsaggelos are with the Department
of Electrical Engineering and Computer Science, Northwestern University, Evanston, IL, 60208 USA. }
\thanks{Manuscript received Feb 3, 2022.}}

%
%

\markboth{Journal of \LaTeX\ Class Files,~Vol.~14, No.~8, August~2015}%
{Shell \MakeLowercase{\textit{et al.}}: Bare Demo of IEEEtran.cls for IEEE Journals}
%



\maketitle

\begin{abstract}
   Dense depth map capture is challenging in existing active sparse illumination based depth acquisition techniques, such as LiDAR. Various techniques have been proposed to estimate a dense depth map based on fusion of the sparse depth map measurement with the RGB image. Recent advances in hardware enable adaptive depth measurements resulting in further improvement of the dense depth map estimation. In this paper, we study the topic of estimating dense depth from depth sampling. The adaptive sparse depth sampling network is jointly trained with a fusion network of an RGB image and sparse depth, to generate optimal adaptive sampling masks. Deep learning based superpixel sampling and soft sampling approximation are applied. We show that such adaptive sampling masks can generalize well to many RGB and sparse depth fusion algorithms under a variety of sampling rates (as low as $0.0625\%$). The proposed adaptive sampling method is fully differentiable and flexible to be trained end-to-end with upstream perception algorithms.
\end{abstract}


%
\IEEEpeerreviewmaketitle

\section{Introduction}
Depth sensing and estimation is important for many applications, such as autonomous driving~\cite{lidarSelfDriving}, augmented reality (AR)~\cite{hololens}, and indoor perception~\cite{han2013enhanced}. 


Based on the principle of operation, we can roughly divide current depth sensors into two categories: (1) Triangulation-based depth sensors (eg., stereo~\cite{kanade1996stereo}) and (2) Time-of-flight (ToF) based depth sensors (including direct ToF LiDAR sensor and indirect ToF cameras~\cite{foix2011lock}). Among these depth sensors, LiDAR has a much longer imaging range (e.g., tens of meters) with a high depth precision (e.g., mm) which enables it to be a competitive depth sensors for numerous emerging commercial applications. LiDAR has been widely used for machine vision applications, such as, navigation in self-driving cars and mobile devices (e.g., LiDAR sensor on Apple iPhone 12).

Most LiDAR sensors illuminate a single point of the object and measure the depth/distance information for that point at a time. LiDAR sensors then rely on raster scanning to generate a full 3D image of the object, which limits the acquisition speed. In order to produce a full 3D image of an object with a reasonable frame rate using mechanical scanning, LiDAR can only provide a sparse scanning with significant inter-sample spacing. This leads to very limited spatial resolution with LiDAR sensors. To increase LiDAR's spatial resolution and acquire more structural information of the scene, other high-spatial-resolution imaging modalities, such as RGB, have been used to be fused with LiDAR's depth images~\cite{van2019sparse,lindell2018single}. 

Traditionally, LiDAR sensors perform raster scanning following a regular grid to produce a uniformly spaced depth map. Recently, researchers explore adaptive illumination/scanning patterns for LiDAR sensors based on the scene and co-optimize the scanning pattern with the multimodal sensor fusion pipeline to further increase performance~\cite{bergman2020deep}. Pittaluga \textit{et al.}~\cite{pittaluga2020towards} implement optimized LiDAR scanning on a real hardware device using a MEMS mirror. They co-optimize the scanning hardware and the fusion pipeline with an RGB sensor to increase LiDAR's performance and achieve higher resolution depth map. Tasneem \textit{et al}. ~\cite{tasneem2020adaptive} utilize adaptive fovea LiDAR scanning to achieve highest angular resolution over regions of interest (ROIs) which can help improve the machine perception accuracy. Yamamoto \textit{et al.}~\cite{yamamoto2018efficient} also propose adaptive LiDAR scanning for efficient and accurate detection on pedestrian with dense scans. With adaptive illumination/sampling, we might also achieve a reasonable depth map with smaller number of samples for post machine perception tasks~\cite{bergman2020deep}, which reduces the sensor bandwidth and may potentially enable higher LiDAR sensor frame rate for those without mechanical scanning~\cite{pittaluga2020towards} where the acquisition time is linearly dependent on the number of samples.

\begin{figure}[h]
\begin{center}
\begin{overpic}[width=0.95\linewidth]{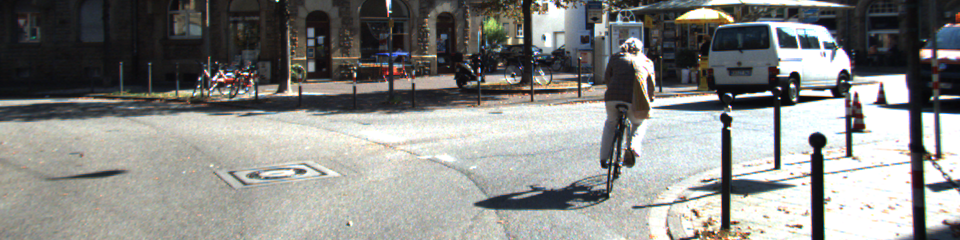}
\put (0,1.2){\color{white}\scriptsize RGB Image}
\end{overpic}
\begin{overpic}[width=0.95\linewidth]{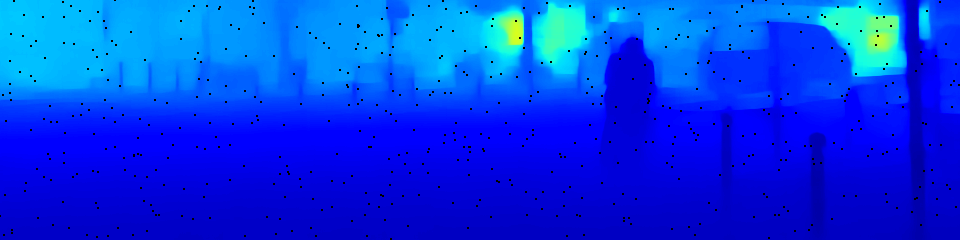}
\put (0,1.2){\color{white}\scriptsize Predicted Depth using Randomly Sampled Depth, RMSE: 2052.5}
\end{overpic}
\begin{overpic}[width=0.95\linewidth]{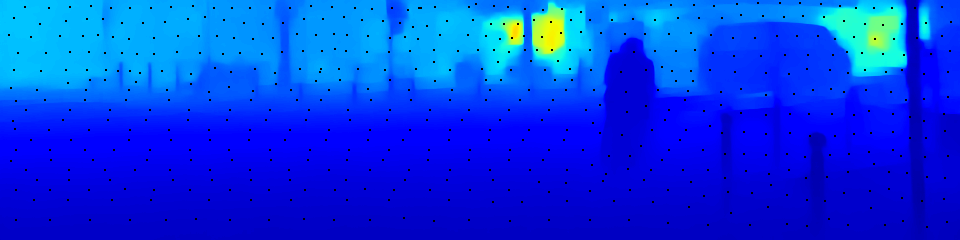}
\put (0,1.2){\color{white}\scriptsize Predicted Depth using Adaptively Sampled Depth, RMSE: 1146.9}
\end{overpic}
\begin{overpic}[width=0.95\linewidth]{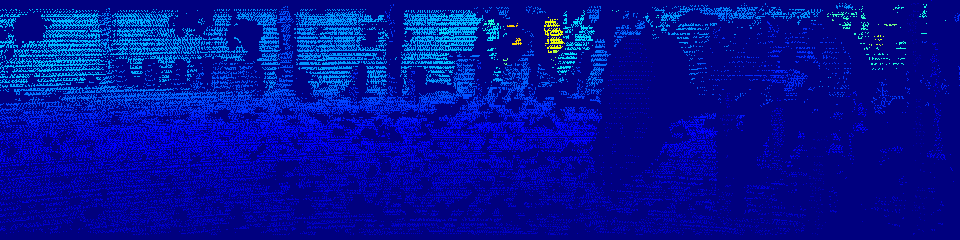}
\put (0,1.2){\color{white}\scriptsize Depth Ground Truth}
\end{overpic}
\end{center}
\caption{LiDAR systems is able to capture accurate sparse depth map (bottom). By reducing the number of samples, we are able to reduce the LiDAR sensor bandwidth which potentially increase the capture frame rate. RGB image (top) can be fused with the captured sparse depth data and estimate a dense depth map. We demonstrate that choosing the sampling location is important to the accuracy of the estimated depth map. Under $0.25\%$ sampling rate (with respect to the RGB image), using the same depth estimation method ~\cite{van2019sparse}, the depth map estimated from the adaptively sampled sparse depth (third row) is more accurate than the depth map estimated from random samples (second row).}
\label{fig:example_rgbd_adaptive_sampling}
\end{figure}

In this paper, we study the topic of adaptive depth sampling and depth map reconstruction. The importance of performing adaptive depth sampling is shown in Figure~\ref{fig:example_rgbd_adaptive_sampling}. First, we formulate the pipeline of joint adaptive depth sampling and depth map estimation. Then, we propose a deep learning (DL) based algorithm for adaptive depth sampling. We show that the proposed adaptive depth sampling algorithm can generalize well to many depth estimation algorithms. Finally, we demonstrate a state-of-the-art depth estimation accuracy compared to other existing algorithms.

Our contribution is summarized as follows:
\begin{itemize}
  \item We propose an adaptive depth sensing framework which benefits from active sparse illumination depth sensors.
  \item We propose a superpixel segmentation based adaptive sampling mask prediction network and a differentiable sampling layer, which translates the estimated sampling locations ($x,y$ coordinates) into a binary sampling mask. We also experimentally show better sampling performance is achieved compared to existing sampling methods.
  \item We demonstrate that the proposed adaptive sampling method can generalize well to many depth estimation algorithms without fine tuning, thus establishing the effectiveness of the proposed sampling method. We also show that the trained adaptive depth sampling networks can generalize across different datasets. According to our knowledge, this is the first study in the literature that performs such generalization tests.
  \item We study the effect of capture time delay between the RGB image and the sampled depth map. We illustrate that the advantage of the proposed adaptive sampling method still holds if we take the temporal registration issue into account.
\end{itemize}

\begin{figure*}[h]
\begin{center}
\includegraphics[width=0.9\linewidth]{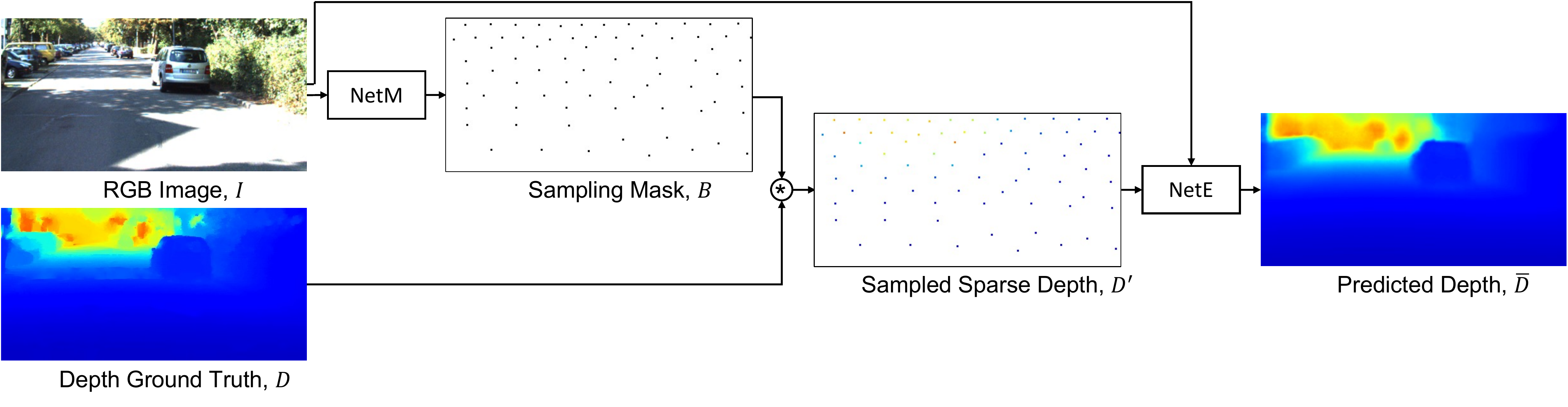}
\end{center}
\caption{The proposed pipeline contains two submodules, adaptive depth sampling Mask computation ($NetM$) and depth Estimation ($NetE$). The binary adaptive sampling mask is generated by $NetM$ based on the RGB image. Then, the LiDAR system samples the scene based on this binary sampling mask and generates the sampled sparse depth map. Finally, both the RGB image and the sampled sparse depth map are input to $NetE$ to estimate a dense depth map.}
\label{fig:proposed_pipeline}
\end{figure*}

\section{Related Work}
In this section, we review work on algorithm-based depth estimation and sampling mask optimization and clarify the relationship of our proposed method to previous work.
\subsection{Depth Estimation}
Given RGB images, early depth prediction methods relied on hand-crafted features and probabilistic graphics models. Karsch \emph{et~al.} \cite{karsch2014depth,karsch2016depth} estimate the depth based on querying an RGBD image database. A Markov random field model is applied in \cite{saxena2005learning} to regress depth from a set of image features. Recently deep learning (DL) and convolutional neural networks (CNNs) have been applied to learn the mapping from single RGB images to dense depth maps \cite{eigen2014depth,eigen2015predicting,laina2016deeper,fu2018deep,qi2018geonet,alhashim2018high,lee2019big, wofk2019fastdepth}. These DL-based approaches achieve state-of-the-art performance because better features are extracted and better mappings are learned from large-scale datasets \cite{silberman2012indoor, geiger2013vision,ros2016synthia}.

Given sparse depth measurements, traditional image filtering and interpolation techniques \cite{ku2018defense} can be applied to reconstruct the dense depth map. Hawe~\cite{hawe2011dense} and Liu~\cite{liu2015depth} study the sparse depth map completion problem from the compressive sensing perspective. DL techniques have also been applied to the sparse depth completion problem. A sparse depth map can either be fed into conventional CNNs \cite{mal2018sparse} or sparsity invariant CNNs \cite{uhrig2017sparsity}. When the sampling rate is low, the sparse depth map completion task is challenging.

If both RGB images and sparse depth measurements are provided, traditional guided filter approaches \cite{levin2004colorization,barron2016fast} can be applied to refine the depth map. Optimization algorithms that promote depth map priors while maintaining fidelity to the observation are proposed in \cite{lu2015sparse,drozdov2016robust,ma2019sparse}. Various DL-based methods have been developed~\cite{mal2018sparse,ma2019self,van2019sparse,chen2018estimating,jaritz2018sparse,wong2020unsupervised,yang2019dense,shivakumar2019dfusenet}. During training and testing, most DL approaches are trained and tested using random or regular grid sampling masks. Because depth completion is an active research area, we do not want to limit our adaptive sampling method to a specific depth estimation method. 

\subsection{Sampling Mask Optimization}
Irregular sampling \cite{cook1986stochastic, piroddi2004analysis, bridson2007fast} is well studied in the computer graphics, image processing and computational imaging literature to achieve good representation of images, and we have witnessed its application in compressive sensing~\cite{wu2019learning}, ghost imaging~\cite{li2020compressive}, wireless imaging~\cite{9076119}, quantitative phase imaging~\cite{kellman2019physics}, Fourier ptychography~\cite{kellman2019data}, and etc. Making the sampling distribution adaptive to the signal can further improve representation performance. Eldar \emph{et~al.} \cite{eldar1997farthest} proposed a farthest point strategy which performs adaptive and progressive sampling of an image. Inspired by the lifting scheme of wavelet generation, several progressive image sampling techniques were proposed~\cite{demaret2006image,rajesh2007fast}. Ramponi \emph{et~al.} \cite{ramponi2001adaptive} applied a measure of the local sample skewness. Lin \emph{et~al.} \cite{lin2015generalized} utilized the generalized Ricci curvature to sample grey scale images as manifolds with density. A kernel construction technique is proposed in \cite{liu2014kernel}. Taimori \emph{et~al.} \cite{taimori2018adaptive} investigated space-frequency-gradient information of image patches for adaptive sampling. 

Specific reconstruction algorithms are needed for each of these irregular or adaptive sampling methods \cite{cook1986stochastic,eldar1997farthest,rajesh2007fast,demaret2006image,ramponi2001adaptive,lin2015generalized,liu2014kernel,taimori2018adaptive} to reconstruct the fully sampled signal. Furthermore, handcrafted features are applied to these sampling methods. Finally, these sampling techniques are all applied to the same modality (RGB or grey scale image). Recently, Dai \emph{et~al.} \cite{dai2019adaptive} applied DL technique to the adaptive sampling problem. The adaptive sampling network is jointly optimized with the image inpainting network. The sampling probability is optimized during training, and binarized during testing. Good performance is demonstrated for X-ray fluorescence (XRF) imaging at a sampling rate as low as $5\%$. Kuznetsov \emph{et~al.} \cite{kuznetsov2018deep} predicted adaptive sampling maps jointly with reconstruction of Monte Carlo (MC) rendered images using DL. A differentiable render simulator with respect to the sampling map was proposed. Huijben \emph{et~al.} \cite{huijben2019deep,huijben2020learning} proposed a task adaptive compressive sensing pipeline. The sampling mask is trained with respect to a specific task and is fixed during imaging. Gumbel-max trick \cite{gumbel1954statistical,jang2016categorical} is applied to make the sampling layer differentiable. 

All of the above DL-based sampling methods predict a per pixel sampling probability~\cite{dai2019adaptive,huijben2019deep,huijben2020learning} or a sampling number~\cite{kuznetsov2018deep}. Good sampling performance has not been demonstrated under extreme low sampling rates $(<1\%)$. Directly enforcing priors on sampling locations is effective when the sampling rate is low. This requires the adaptive sampling network to predict sampling locations ($(x,y)$ coordinates) directly and the sampling process to be differentiable. For the RGB and sparse depth adaptive sampling task, Wolff \emph{et~al.} \cite{wolff2020super} use the SLIC superpixel technique~\cite{achanta2012slic} to segment the RGB image and sample the depth map at the center of mass of each superpixel. A bilateral filtering based reconstruction algorithm is proposed to reconstruct the depth map. A spatial distribution prior is implicitly enforced by superpixel segmentation, resulting in good sampling performance under low sampling rates. The sampling and reconstruction methods are not optimized jointly, leaving room for improvement. In this paper, we show that jointly training recent DL-based superpixel sampling networks \cite{jampani2018superpixel,wolff2020super} and depth estimation networks \cite{cook1986stochastic,eldar1997farthest,rajesh2007fast,demaret2006image,ramponi2001adaptive,lin2015generalized,liu2014kernel,taimori2018adaptive} can be adapted to the problem of dense depth map estimation from sparse LiDAR data with improved accuracy. Bergman \emph{et~al.} \cite{bergman2020deep} warp a uniform sampling grid to generate the adaptive sampling mask. The warping vectors are computed utilizing DL-based optical flow estimated from the RGB image. A spatial distribution prior is enforced by the initial uniform sampling grid. End-to-end optimization of the sampling and depth estimation networks is performed and good depth reconstruction is obtained under low sampling rates. In the pipeline of~\cite{bergman2020deep}, there are 4 sub-networks, 2 for sampling and the other 2 for depth estimation. They are jointly trained but only the final depth estimation results are demonstrated. The whole pipeline is bulky and expensive. More importantly, it is hard to assess if the improvement on depth estimation comes from the sampling part or the depth estimation part of the pipeline. In this paper, we decouple these two parts and study each individual module to better understand their contribution towards the final depth estimate. Finally, a bilinear sampling kernel is applied in~\cite{bergman2020deep} to make the optimization of the sampling locations differentiable. In contrast, we propose a novel differentiable relaxation of the sampling procedure and show its advantages over the bilinear sampling kernel.

\section{Method}

\subsection{Problem Formulation}
As shown in Figure~\ref{fig:proposed_pipeline}, the input RGB image is denoted by $I$. The mask generation network $NetM$ produces a binary sampling mask $B=NetM(I,c)$, where $c\in [0, 1]$ is the predefined sampling rate. Elements in $B$ equal to $1$ correspond to sampling locations and $0$ to non-sampling location. The LiDAR system samples depth according to $B$ and produces the measured sparse depth map $D^{\prime}$. In synthetic experiments, if the ground truth depth map $D$ is given, the measured sparse depth map $D^{\prime}$ is obtained according to

\begin{equation}
\label{e:e1}
D^{\prime} = D \odot B = D \odot NetM(I, c),
\end{equation} where $\odot$ is the element-wise product operation. The reconstructed depth map $\bar{D}$ is obtained by the depth estimation network $NetE$, that is,

\begin{equation}
\label{e:e2}
\bar{D} = NetE(I, D^{\prime}) = NetE(I, D \odot NetM(I,c)).
\end{equation}

The overall adaptive depth sensing and depth estimation pipeline is shown in Figure~\ref{fig:proposed_pipeline}. End-to-end training can be applied on $NetM$ and $NetE$ jointly. The adaptive depth sampling strategy is learned by $NetM$, while $NetE$ estimates the final dense depth map. An informative sampling mask is beneficial to depth estimation algorithms in general, not just to $NetE$. Given a limitted depth sampling budget and an RGB image, we want to sample depth value on the ambiguous regions in a balanced way. During testing, we can replace the inpainting network $NetE$ with other depth estimation algorithms. Network architectures and training details of $NetE$ and $NetM$ are discussed in the following subsections.

\subsection{Depth Estimation Network $NetE$}

We use the network architecture in~\cite{mal2018sparse} for the depth estimation network. The network is an encoder-decoder pipeline. The encoder takes a concatenated $I$ and $D^{\prime}$ as input ($4$ channels) and encodes them into latent features. The decoder takes the low spatial resolution feature representation and outputs the restored depth map $\bar{D} = NetE(I, D^{\prime})$. Readers can refer to~\cite{mal2018sparse} for the detailed architecture of $NetE$.

Because method \cite{mal2018sparse} is differentiable with respect to $D^{\prime}$ (unlike \cite{chen2018estimating}) and its network architecture is standard without customized fusion modules~\cite{jaritz2018sparse,van2019sparse,chen2018estimating}, we choose it as $NetE$ and jointly train $NetM$ with it according to Figure~\ref{fig:proposed_pipeline}. We found out that the trained $NetM$ can generalize well to other depth estimation methods during testing. 

\subsection{Sampling Mask Generation Network $NetM$}
\label{sec:netm}
\begin{figure}[b]
\begin{center}
\includegraphics[width=0.95\linewidth]{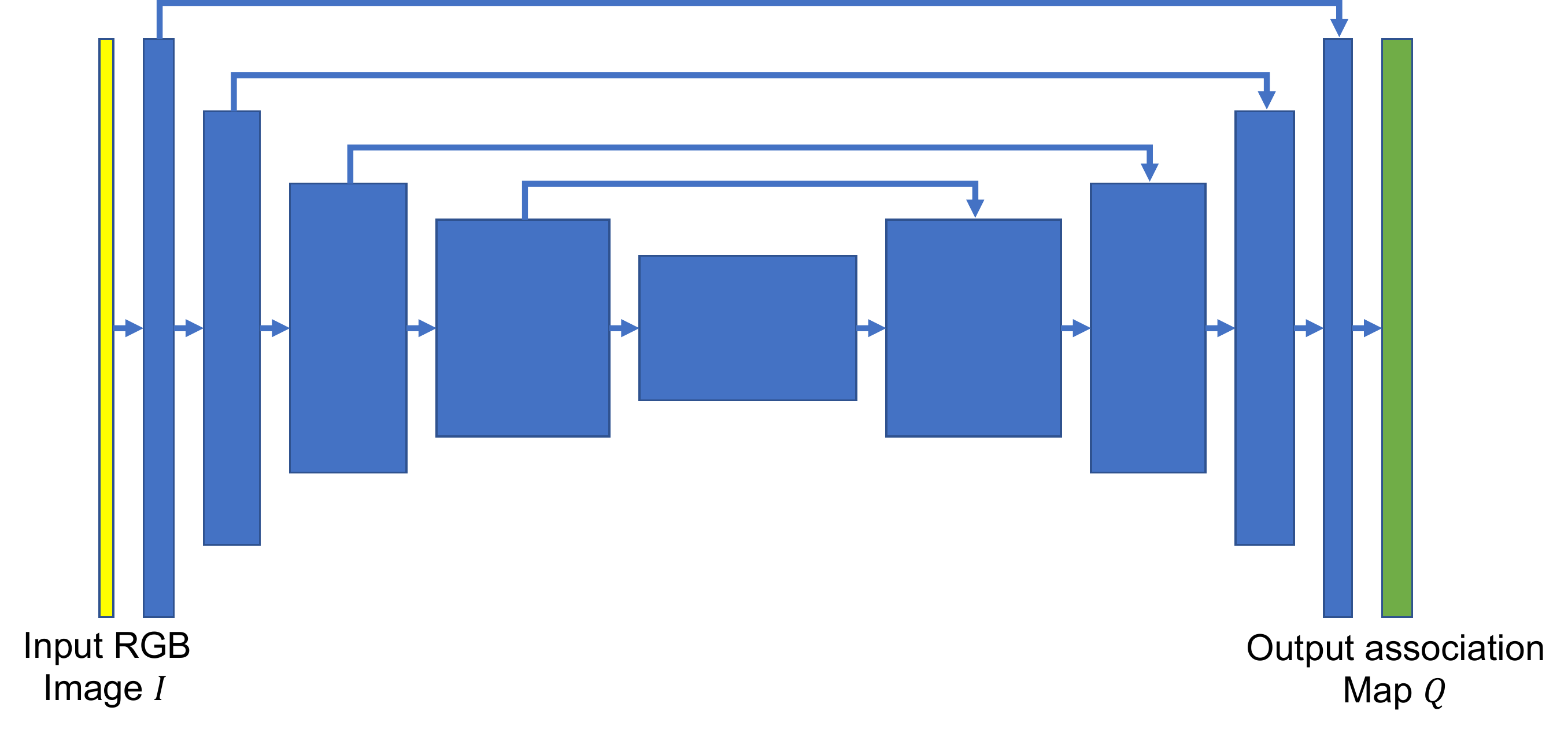}
\end{center}
\caption{Superpixel FCN~\cite{yang2020superpixel}'s encoder-decoder network architecture. }
\label{fig:super_pixel_u_net}
\end{figure}

Existing irregular sampling techniques~\cite{eldar1997farthest, bridson2007fast} and adaptive depth sampling methods~\cite{bergman2020deep,wolff2020super} explicitly or implicitly make sampling points evenly distributed spatially. Such prior is important when the sampling rate is low. Inspired by the SLIC superpixel~\cite{achanta2012slic} based adaptive sampling method~\cite{wolff2020super}, we propose to utilize recent DL-based superpixel networks~\cite{yang2020superpixel,jampani2018superpixel} as $NetM$. As demonstrated in Figure~\ref{fig:proposed_pipeline}, $NetM$ adapts to the task of depth sampling after being jointly trained with $NetE$.

Superpixel with fully convolutional networks (FCN)~\cite{yang2020superpixel} is one of the DL-based superpixel techniques. It predicts the pixel association map $Q$ given an RGB image $I$. Its encoder-decoder network architecture is shown in Figure~\ref{fig:super_pixel_u_net}. Similar to the SLIC superpixel method~\cite{achanta2012slic}, a combined loss that enforces similarity property of pixels inside one superpixel and spatial compactness is applied. Readers can refer to~\cite{yang2020superpixel} for more details.

Given an RGB image $I$ with spatial dimensions $(H, W)$, under the desired depth sampling rate $c$, we have $N_p=H \cdot W$ pixels and $N_s=c \cdot H \cdot W$ superpixels. The sampled depth location is the weighted mass center of each superpixel. We denote the subset of pixels as $\mathcal{P}=\{\mathcal{P}_0, ...,\mathcal{P}_{N_s-1}\}$, where $\mathcal{P}_i$ is a set of pixels associated with superpixel $i$. Pixel $p$'s CIELAB color property and $(x,y)$ coordinates are denoted by $\mathbf{f}(p) \in \mathbb{R}^{3}$ and $\mathbf{c}(p)\in \mathbb{R}^{2}$, respectively. CIELAB color space is used here as we follow the FCN~\cite{yang2020superpixel} and SLIC superpixel~\cite{achanta2012slic} setup. The loss function is given by

\begin{equation}
\label{e:e3}
\mathcal{L}_{SLIC}(\mathbf{f}, Q)=\sum_{p\in \mathcal{P}}\Vert \mathbf{f}(p) - \mathbf{f}^{\prime}(p)\Vert_2+m\Vert \mathbf{c}(p) - \mathbf{c}^{\prime}(p)\Vert_2.
\end{equation} 
Here we have
\begin{subequations}
\label{e:e4}
\begin{align}
\mathbf{u}_s = \frac{\sum_{p\in \mathcal{P}_s}\mathbf{f}(p)q_{s}(p)}{\sum_{p\in \mathcal{P}_s}q_{s}(p)}, &  &\mathbf{l}_s  = \frac{\sum_{p\in \mathcal{P}_s}\mathbf{c}(p)q_{s}(p)}{\sum_{p\in \mathcal{P}_s}q_{s}(p)},\\
\mathbf{f}^{\prime}(p) = \sum_{s\in \mathcal{N}_p}\mathbf{u}_{s} q_{s}(p), & &  \mathbf{c}^{\prime}(p) = \sum_{s\in \mathcal{N}_p}\mathbf{l}_{s} q_{s}(p),
\end{align}
\end{subequations} where $m$ is a weight balancing term between the CIELAB color similarity and spatial compactness, $\mathcal{N}_p$ is the set of superpixels surrounding $\mathbf{p}$, $q_s(p)$ is the probability of a pixel $p$ being associated with superpixel $s$ and is derived from the associate map $Q$, $\mathbf{u}_s\in \mathbb{R}^{3}$ and $\mathbf{l}_s\in \mathbb{R}^{2}$ are the color property and locations of superpixel $s$, $\mathbf{f}^{\prime}(p)\in \mathbb{R}^{3}$ and $\mathbf{c}^{\prime}(p)\in \mathbb{R}^{2}$ are respectively the reconstructed color property and location of pixel $p$. 

\subsection{Soft Sampling Approximation}
\label{sec:soft_sampling_approximation}

Defined in Equation~\ref{e:e4}(a), we denote the collection of $\mathbf{l}_s$, $s=0,...,N_s-1$ , as $S$. Depth values at locations $S$ would be measured during the depth sampling. In order to train $NetM$ and $NetE$ jointly, the sampling operation $g$, which computes the sampled sparse depth map $D^{\prime}$ from depth ground truth $D$ and sampling location $S$, $D^{\prime}=g(D, S)$, needs to be differentiable with respect to $S$. Unfortunately, such sampling operation $g$ is not differentiable in practice. Bergman \emph{et~al.}~\cite{bergman2020deep} apply a bilinear sampling kernel to differentiably correlate $S$ and $D^{\prime}$. The computed gradients rely on the $2\times 2$ local structure of the ground truth depth map $D$. The computed gradients are not stable when the sampling location is sparse. Thus limited sampling performance is obtained. We propose a soft sampling approximation (SSA) strategy during training. SSA utilizes a larger window size compared to the bilinear kernel and achieves better sampling performance.

\begin{figure}[!t]
\begin{center}
\subfigure[]{\includegraphics[width=0.32\linewidth]{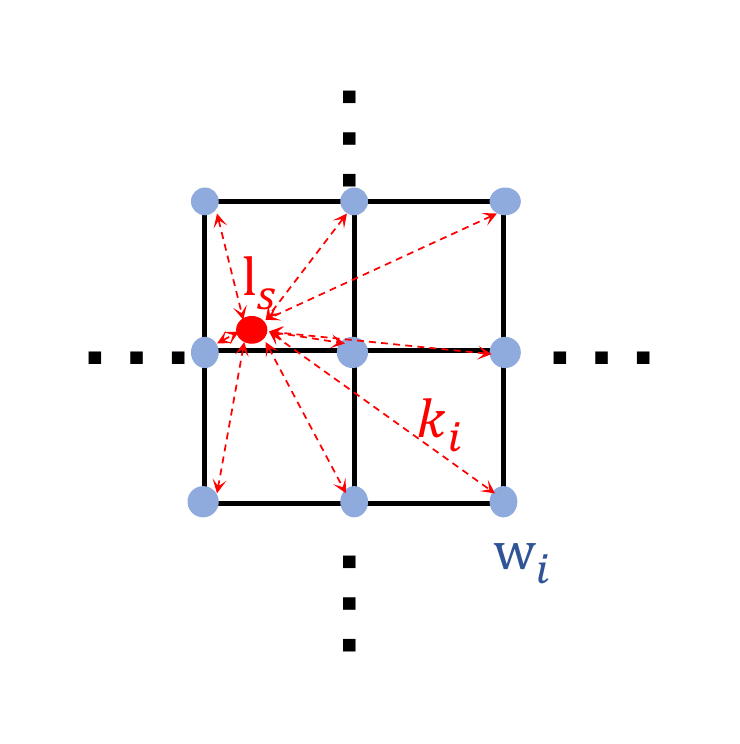}} 
\subfigure[]{\includegraphics[width=0.32\linewidth]{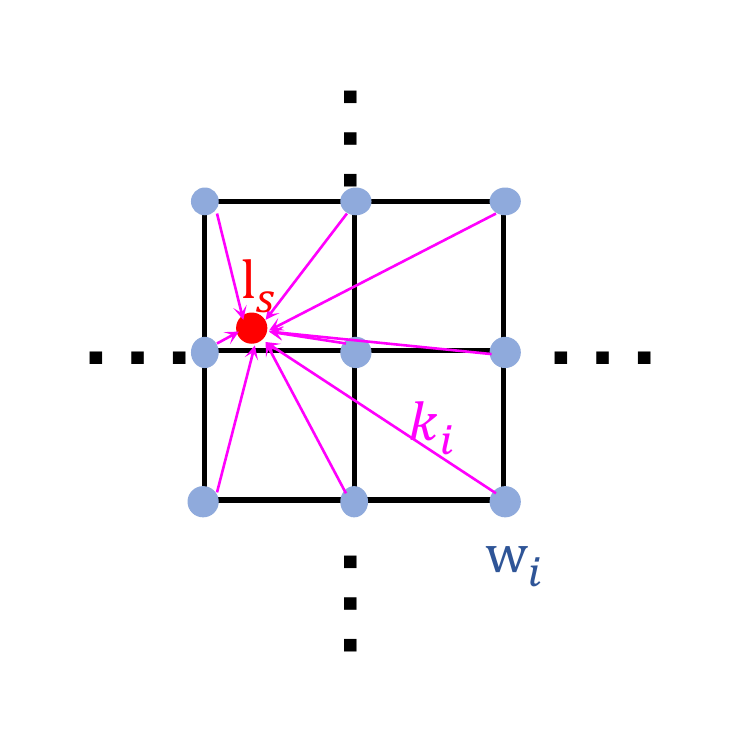}} 
\subfigure[]{\includegraphics[width=0.32\linewidth]{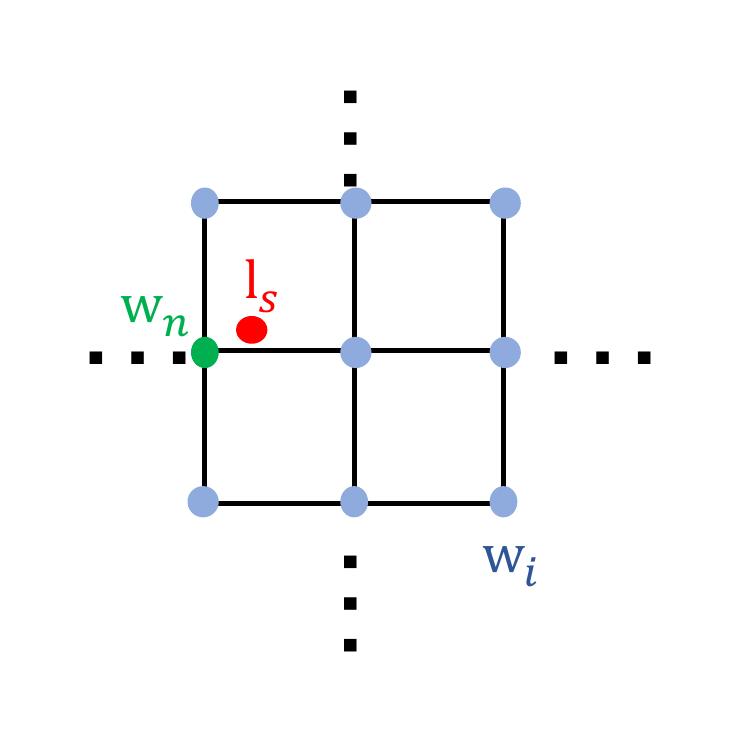}} 
\end{center}
\caption{Illustration of the sampling approximation. (a) We find a local window $W$ of $\mathbf{l}_s$ and compute distance $\rho_{i}$. (b) We represent $\mathbf{l}_s$'s depth value $d_s$ as a linear combination of local window $W$'s depth values. (c) During testing, we sample the depth value at the nearest neighbour $\mathbf{w}_n$ of $\mathbf{l}_s$.}
\label{fig:soft_sampling_trick}
\end{figure}

As shown in Figure~\ref{fig:soft_sampling_trick}, during training, given a sampling location $\mathbf{l}_s\in S$, we find a local $h \times w$ window $W$ around $\mathbf{l}_s$. The depth value $d_s$ at $\mathbf{l}_s$ is a weighted average of the depth values in $W$,

\begin{equation}
\label{e:e5}
d_s=\sum_{i\in \mathcal{N}_{W}} k_i d_i,
\end{equation} where $\mathcal{N}_{W}$ includes the indices of all pixels in $W$, $\mathbf{w}_i$ is the $i^{th}$ pixel's location in $W$, $d_i$ is the depth value of $\mathbf{w}_i$, the weights $k_i$ are computed according to the Euclidean distance $\rho_i$ between $\mathbf{l}_s$ and $\mathbf{w}_i$, scaled by a temperature parameter $t$,

\begin{equation}
\label{e:e6}
k_i=\frac{e^{-\rho_i^2/t^2}}{\sum_{j\in \mathcal{N}_{W}}e^{-\rho_j^2/t^2}}.
\end{equation}

When the temperature parameter $t\rightarrow 0$, the sampled depth value $d_s$ is equal to the depth value $d_n$ of the nearest pixel $\mathbf{w}_n$. When $t$ is large, the soft sampled depth value $d_s$ is different from $d_n$. We gradually reduce $t$ during the training process. During testing, we find the nearest neighbor pixel $\mathbf{w}_n$ of $\mathbf{l}_s$ and sample the depth value $d_n$ at $\mathbf{w}_n$.

\subsection{Training Procedures}
\label{sec:training_procedures}
Given the training dataset consisting of the aligned RGB image $I$ and the ground truth depth map $D$, we first train $NetE$ by minimizing the depth loss,
\begin{equation}
\label{e:e7}
\mathcal{L}_{depth}=\Vert D - NetE(I,D^{\prime})\Vert_2,
\end{equation} where $D^{\prime}$ is obtained by applying a random sampling mask on $D$ with sampling rate $c$.

Then we initialize the superpixel network $NetM$ using the RGB image $I$. $\mathcal{L}_{SLIC}$ is minimized according to Equation~\ref{e:e3}. The initialized $NetM$ approximates the SLIC superpixel segmentation on RGB image. If we sample the depth value on $\mathbf{l}_s$ of each superpixel, the sampling pattern would be similar to ~\cite{wolff2020super}.

Finally, we freeze $NetE$ and train $NetM$ in Figure~\ref{fig:proposed_pipeline} by minimizing
\begin{equation}
\label{e:e8}
\mathcal{L}=\mathcal{L}_{depth} + q\cdot\mathcal{L}_{SLIC},
\end{equation} where $q$ is the weighting terms of $\mathcal{L}_{SLIC}$. The SSA trick shown in Figure~\ref{fig:soft_sampling_trick} is applied and the temperature parameter $t$ gradually decreases during training.

\begin{table*}[h]
\scriptsize
\begin{tabular}{c | c | c | c | c | c | c | c | c | c | c | c | c } \hline \hline
 & \multicolumn{12}{c}{MAE (mm)} \\  \cline{2-13}
 & \multicolumn{4}{c|}{c=1\%} & \multicolumn{4}{c|}{c=0.25\%} & \multicolumn{4}{c}{c=0.0625\%} \\  \cline{2-13}
 & FusionNet & SSNet & $NetE$  & Colorization  & FusionNet & SSNet & $NetE$ & Colorization  & FusionNet & SSNet &  $NetE$ & Colorization  \\ \hline
Random & 324.8 & 466.6 & 425.7 & 764.6 & 488.6 & 654.9 & 557.1 & 1390.7 & 798.5 & 1021.1 & 779.4 & 2517.5 \\ \hline
Uniform Grid & 301.4 & 450.2 & 398.2 & 694.6 & 439.0 & 598.0 & 516.4 & 1257.3 & 692.5 & 843.2 & 715.3 & 2247.3 \\ \hline
Poisson~\cite{bridson2007fast} & 324.1 & 451.8 & 409.6 & 711.5 & 455.8 & 621.2 & 537.5 & 1314.1 & 736.2 & 901.0 & 743.1 & 2428.4 \\ \hline
SPS~\cite{wolff2020super} & 297.2 & 439.9 & 388.1 & \textit{654.3} & 436.9 & 594.2 & 507.8 & \textit{1197.2} & 713.8 & 865.2 & 724.5 & \textbf{2175.9}\\ \hline
DAL~\cite{bergman2020deep} & 295.8 & 447.4 & 390.1 & 683.4 & 432.5 & 599.1 & 504.8 & 1239.7 & 694.4 & 838.4 & 710.2 & 2230.7 \\ \hline
FCN~\cite{yang2020superpixel} & 298.2 & 440.8 & 390.0 & 672.5 & 426.8 & \textit{587.4} & \textit{498.5} & 1202.7 & 683.8 & 833.6 & 698.2 & 2227.6 \\ \hline
$NetM-NYU$ & \textit{291.3} & \textit{435.9} & \textit{382.0} & 662.7 & \textit{425.9} & 590.1 & 499.2 & 1212.4 & \textit{657.4} & \textit{799.7} & \textit{678.9} & \textit{2221.0} \\ \hline
$NetM$ & \textbf{285.0} & \textbf{423.1} & \textbf{380.1} & \textbf{656.2} & \textbf{404.3} & \textbf{562.2} & \textbf{477.5} & \textbf{1189.2} & \textbf{634.9} & \textbf{778.0} & \textbf{652.2} & 2265.8\\ \hline\hline

 & \multicolumn{12}{c}{RMSE (mm)} \\ \cline{2-13}
 & \multicolumn{4}{c|}{c=1\%} & \multicolumn{4}{c|}{c=0.25\%} & \multicolumn{4}{c}{c=0.0625\%} \\ \cline{2-13}
 & FusionNet & SSNet & $NetE$  & Colorization  & FusionNet & SSNet & $NetE$ & Colorization  & FusionNet & SSNet &  $NetE$ & Colorization  \\ \hline
Random & 1060.0	& 1221.6 & 1294.8 & 1984.4 & 1476.0 & 1709.9 & 1704.3 & 3087.3 & 2135.6 & 2505.6 & 2262.61 & 4749.8 \\ \hline
Uniform Grid & 988.4 & 1139.2 & 1207.8 & 1840.9 & 1359.1 & 1570.2 & 1566.0 & 2854.2 & 1946.4 & 2132.5 & 2101.7 & 4315.9 \\ \hline
Poisson~\cite{bridson2007fast} & 1010.1 & 1140.2 & 1193.3 & 1844.8 & 1375.6 & 1589.1 & 1596.8 & 2897.3 & 2013.1 & 2256.0 & 2151.4 & 4508.6 \\ \hline
SPS~\cite{wolff2020super} & 1039.1 & 1124.3 & 1160.5 & \textit{1742.9} & 1360.1 & 1559.7 & 1553.1 & \textit{2718.1} & 1993.8 & 2215.0 & 2141.5 & 4161.1 \\ \hline
DAL~\cite{bergman2020deep} & 969.9 & 1115.1 & 1177.5 & 1784.3 & 1336.1 & 1548.1 & 1532.1 & 2772.6 & 1937.6 & 2128.3 & 2085.7 & 4242.3 \\ \hline
FCN~\cite{yang2020superpixel} & 982.8 & 1123.6 & 1165.1 & 1778.7 & 1324.1 & \textit{1530.4} & 1517.1 & 2728.8 & 1893.9 & 2103.3 & 2046.2 & 4188.5\\ \hline
$NetM-NYU$ & \textit{957.2} & \textit{1095.1} & \textit{1139.2} & 1744.3 & \textit{1322.0} & 1532.7 & \textit{1514.7} & 2743.1 & \textit{1849.6} & \textit{2022.1} & \textit{2010.6} & \textit{4031.8} \\ \hline
$NetM$ & \textbf{939.4} & \textbf{1074.9} & \textbf{1131.3} & \textbf{1725.9} & \textbf{1239.7} & \textbf{1436.8} & \textbf{1422.4} & \textbf{2584.5} & \textbf{1732.4} & \textbf{1930.5} & \textbf{1896.7} & \textbf{3972.9}\\ \hline \hline

\end{tabular}

\smallskip

\caption{Depth sampling and estimation results on KITTI depth completion dataset. Random, Poisson~\cite{bridson2007fast}, Uniform Grid, SPS~\cite{wolff2020super}, DAL~\cite{bergman2020deep}, FCN~\cite{yang2020superpixel} and proposed $NetM$ sampling strategies are compared utilizing $NetE$~\cite{mal2018sparse}, FusionNet~\cite{van2019sparse}, SSNet~\cite{ma2019self}, and Colorization~\cite{levin2004colorization} depth estimation algorithms. MAE and RMSE metrics are reported. Best results are shown in bold. Second best results are shown in italic. The results shown are averaged over a set of $3426$ test frames.}
\label{table:depth_sampling_estimation_table}
\end{table*}

\begin{table*}[h]
\scriptsize
\begin{tabular}{c | c | c | c | c | c | c | c | c | c | c | c | c } \hline \hline
 & \multicolumn{12}{c}{MAE (mm)} \\  \cline{2-13}
 & \multicolumn{4}{c|}{c=1\%} & \multicolumn{4}{c|}{c=0.25\%} & \multicolumn{4}{c}{c=0.0625\%} \\  \cline{2-13}
 & FusionNet & SSNet & $NetE$  & Colorization  & FusionNet & SSNet & $NetE$ & Colorization  & FusionNet & SSNet &  $NetE$ & Colorization  \\ \hline
Random & 39.27 & 59.74 & 94.62 & 74.14 & 73.41 & 104.82 & 99.63 & 146.08 & 146.51 & 186.23 & 151.33 & 283.85 \\ \hline
Uniform Grid & 36.39 & 56.52 & 89.09 & 65.45 & 65.35 & 98.79 & 91.95 & 128.90 & 123.90 & 151.03 & 134.33 & 244.17 \\ \hline
Poisson~\cite{bridson2007fast} & 35.84 & 54.09 & 89.55 & 66.17 & 65.54 & 96.91 & 94.99 & 135.73 & 140.44 & 177.40 & 151.41 & 298.46\\ \hline
SPS~\cite{wolff2020super} & \textit{34.25} & \textit{52.79} & \textit{86.50} & \textbf{59.47} & \textit{61.43} & \textbf{92.68} & \textit{88.95} & \textbf{119.76} & 126.06 & \textit{153.82} & \textit{134.28} & \textbf{238.34}\\ \hline
DAL~\cite{bergman2020deep} & 36.40 & 55.64 & 86.69 & 65.44 & 65.56 & 98.41 & 92.17 & 129.13 & 127.15 & 154.91 & 135.54 & 244.15 \\ \hline
FCN~\cite{yang2020superpixel} & 35.20 & 53.66 & 87.43 & 63.27 & 62.62 & 94.52 & 90.03 & 124.20 & 123.60 & \textbf{150.87} & 134.52 & \textit{239.31}\\ \hline
$NetM-KITTI$ & 34.78 & 53.63 & 87.25 & 64.22 & 62.96 & 94.36 & 90.02 & 128.73 & \textit{123.44} & 154.63 & 134.59 & 252.81 \\ \hline
$NetM$ & \textbf{34.08} & \textbf{52.64} & \textbf{86.38} & \textit{62.54} & \textbf{61.31} & \textit{93.18} & \textbf{88.85} & \textit{124.05} & \textbf{123.10} & 154.41 & \textbf{133.64} & 252.03\\ \hline\hline

 & \multicolumn{12}{c}{RMSE (mm)} \\ \cline{2-13}
 & \multicolumn{4}{c|}{c=1\%} & \multicolumn{4}{c|}{c=0.25\%} & \multicolumn{4}{c}{c=0.0625\%} \\ \cline{2-13}
 & FusionNet & SSNet & $NetE$  & Colorization  & FusionNet & SSNet & $NetE$ & Colorization  & FusionNet & SSNet &  $NetE$ & Colorization  \\ \hline
Random & 98.05 & 116.79 & 167.91 & 150.76 & 156.80 & 182.50 & 192.17 & 249.83 & 255.80 & 303.42 & 269.22 & 425.56\\ \hline
Uniform Grid & 94.00 & 111.01 & 152.33 & 138.36 & 143.39 & 169.00 &	178.29 & 224.29 & 228.21 & 249.13 & 245.00 & 370.02 \\ \hline
Poisson~\cite{bridson2007fast} & 91.07 & 106.34 & 152.11 & 137.53 & 142.22 & 165.17 & 180.39 & 230.84 & 245.22 & 278.14 & 259.86 & 436.13 \\ \hline
SPS~\cite{wolff2020super} & \textit{88.19} & \textbf{102.89} & \textit{144.55} & \textbf{126.23} & \textit{136.01} & \textbf{158.49} & \textit{169.80} & \textbf{209.53} & 228.45 & 252.40 & 244.14 & \textbf{357.96} \\ \hline
DAL~\cite{bergman2020deep} & 93.87 & 110.24 & 149.41 & 138.13 & 143.56 & 168.79 & 177.96 & 225.04 & 231.91 & 254.32 & 246.99 & 368.31 \\ \hline
FCN~\cite{yang2020superpixel} & 90.71 & 106.21 & 147.56 & 132.95 & 137.73 & 161.93 & 173.05 & 217.14 & 226.15 & \textit{247.40} & 242.05 & \textit{361.41}\\ \hline
$NetM-KITTI$ & 89.52 & 105.57 & 146.37 & 134.58 & 138.00 & 161.05 & 172.81 & 221.45 & \textit{223.65} & 248.00 & \textit{239.59} & 374.80 \\ \hline
$NetM$ & \textbf{87.61} & \textit{103.37} & \textbf{143.94} & \textit{130.60} & \textbf{134.60} & \textit{158.69} & \textbf{169.41} & \textit{214.63} & \textbf{222.59} & \textbf{247.05} & \textbf{238.97} & 375.03\\ \hline \hline

\end{tabular}

\smallskip

\caption{Depth sampling and estimation results on NYU-Depth-V2 dataset. Random, Poisson~\cite{bridson2007fast}, Uniform Grid, SPS~\cite{wolff2020super}, DAL~\cite{bergman2020deep}, FCN~\cite{yang2020superpixel} and proposed $NetM$ sampling strategies are compared utilizing $NetE$~\cite{mal2018sparse}, FusionNet~\cite{van2019sparse}, SSNet~\cite{ma2019self}, and Colorization~\cite{levin2004colorization} depth estimation algorithms. MAE and RMSE metrics are reported. Best results are shown in bold. Second best results are shown in italic. The results shown are averaged over a set of $654$ test frames.}
\label{table:depth_sampling_estimation_table_nyu}
\end{table*}

We fix $NetE$ when training $NetM$. Optimizing $NetE$ and $NetM$ simultaneously would obtain $better$ depth reconstruction accuracy~\cite{bergman2020deep}. However, similarly to~\cite{dai2019adaptive}, we would utilize other depth estimation methods than $NetE$ during testing. We want to make the adaptive depth sampling mask be general and applicable to many depth estimation algorithms. 

\section{Experimental Results}
\subsection{Implementation Details}

We use both the KITTI depth completion dataset~\cite{uhrig2017sparsity} and the NYU-Depth-V2 dataset~\cite{silberman2012indoor} for our experiments. The KITTI depth completion dataset consists of aligned ground truth depth maps (from LiDAR sensor) and RGB images. The original KITTI training and validation set split is applied. There are $42949$ and $3426$ frames in the training and testing sets, respectively. We only use the bottom center crop $240\times 960$ of the images because the LiDAR sensor has no measurements at the upper part of the images. The NYU-Depth-V2 dataset consists of RGB and depth images captured by a Microsoft Kinect. $48004$ synchronized RGB-depth image pairs are used for training. $654$ RGB-Depth image pairs from the small labeled test dataset are used for testing. Following ~\cite{mal2018sparse}, the original frames of resolution $480\times 640$ are down sampled to half resolution, producing a final resolution of $240\times 320$.

For the KITTI depth completion dataset, the ground truth depth maps are not dense because they are measured by a velodyne LiDAR device. In order to perform adaptive depth sampling, we need dense depth maps to sample from. Similarly to~\cite{bergman2020deep}, a traditional image inpainting algorithm~\cite{levin2004colorization} is applied to densify the depth ground truth. During evaluation, we compare the estimated dense depth maps to the original sparse ground truth depth maps. For the NYU-Depth-V2 dataset, the raw depth values are projected onto the synchronized RGB images and inpainted using a bilateral filter method available in the official toolbox.

\begin{figure*}[h]
\tiny
\begin{center}
\setlength{\tabcolsep}{0.1pt}
\begin{tabular}{c c c c }

\begin{overpic}[width=0.25\linewidth]{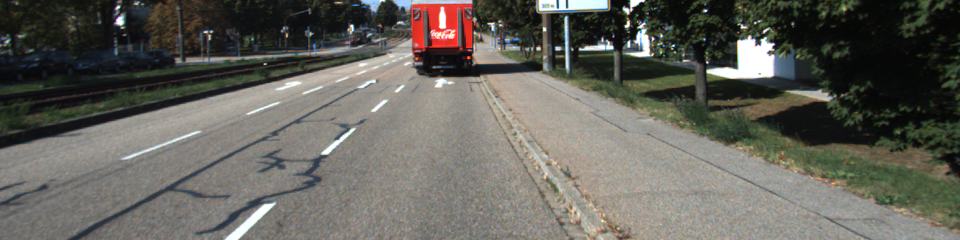}
\put (0,1.2){\color{white}\tiny RGB Image}
\end{overpic}
&
\begin{overpic}[width=0.25\linewidth]{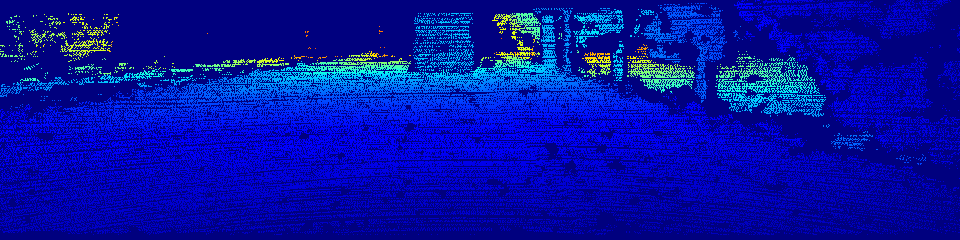}
\put (0,1.2){\color{white}\tiny Depth Ground Truth}
\end{overpic}
&
& \\
\begin{overpic}[width=0.25\linewidth]{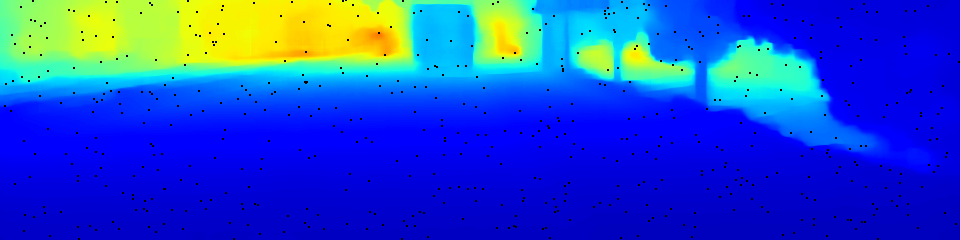}
\put (0,4.2){\color{white}\tiny RMSE: 1612.2}
\put (0,1.2){\color{white}\tiny Random Sampling + FusionNet Reconstruction}
\end{overpic}
&
\begin{overpic}[width=0.25\linewidth]{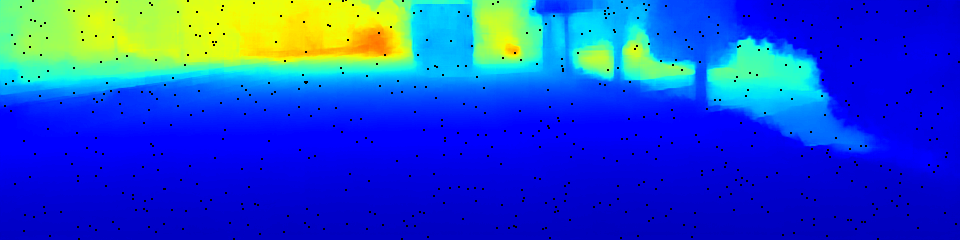}
\put (0,4.2){\color{white}\tiny RMSE: 1913.8}
\put (0,1.2){\color{white}\tiny Random Sampling + SSNet Reconstruction}
\end{overpic}
&
\begin{overpic}[width=0.25\linewidth]{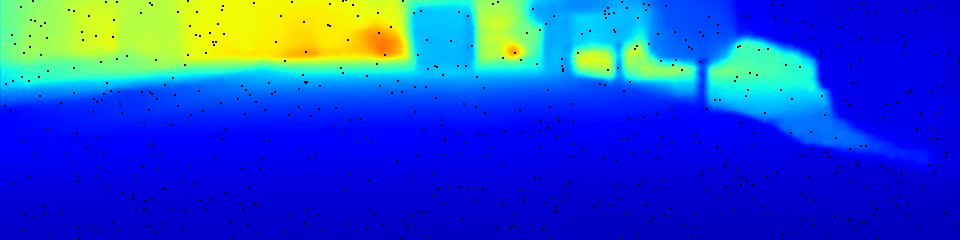}
\put (0,4.2){\color{white}\tiny RMSE: 1852.0}
\put (0,1.2){\color{white}\tiny Random Sampling + $NetE$ Reconstruction}
\end{overpic}
&
\begin{overpic}[width=0.25\linewidth]{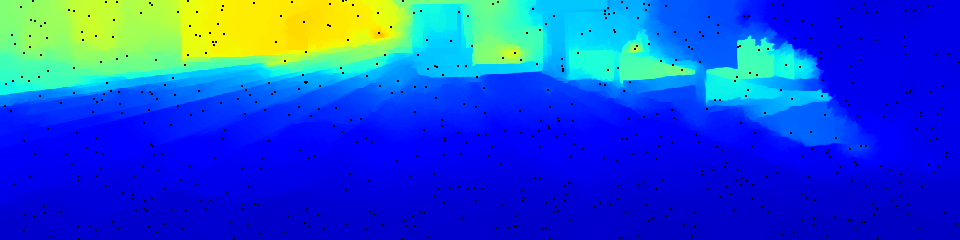}
\put (0,4.2){\color{white}\tiny RMSE: 3604.3}
\put (0,1.2){\color{white}\tiny Random Sampling + Colorization Reconstruction}
\end{overpic}
\\
\begin{overpic}[width=0.25\linewidth]{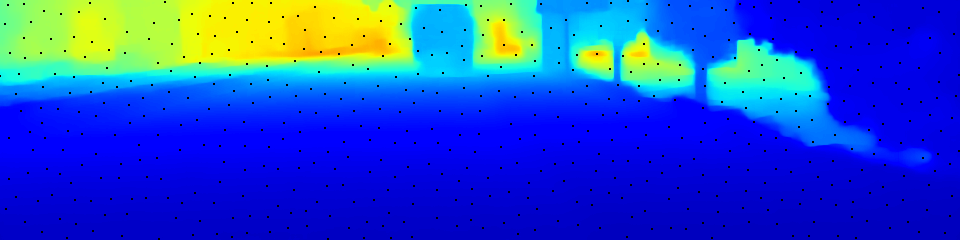}
\put (0,4.2){\color{white}\tiny RMSE: 1617.8}
\put (0,1.2){\color{white}\tiny Poisson Sampling + FusionNet Reconstruction}
\end{overpic}
&
\begin{overpic}[width=0.25\linewidth]{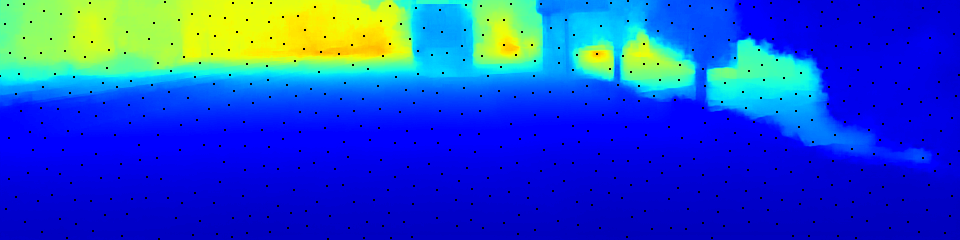}
\put (0,4.2){\color{white}\tiny RMSE: 1873.0}
\put (0,1.2){\color{white}\tiny Poisson Sampling + SSNet Reconstruction}
\end{overpic}
&
\begin{overpic}[width=0.25\linewidth]{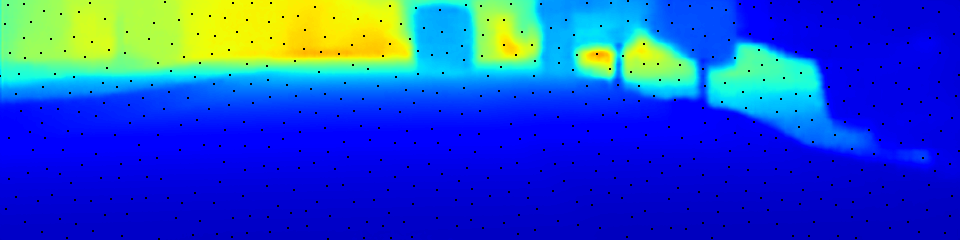}
\put (0,4.2){\color{white}\tiny RMSE: 1965.0}
\put (0,1.2){\color{white}\tiny Poisson Sampling + $NetE$ Reconstruction}
\end{overpic}
&
\begin{overpic}[width=0.25\linewidth]{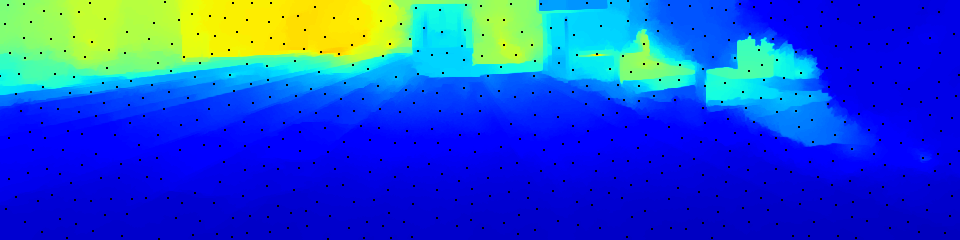}
\put (0,4.2){\color{white}\tiny RMSE: 3260.5}
\put (0,1.2){\color{white}\tiny Poisson Sampling + Colorization Reconstruction}
\end{overpic}
\\
\begin{overpic}[width=0.25\linewidth]{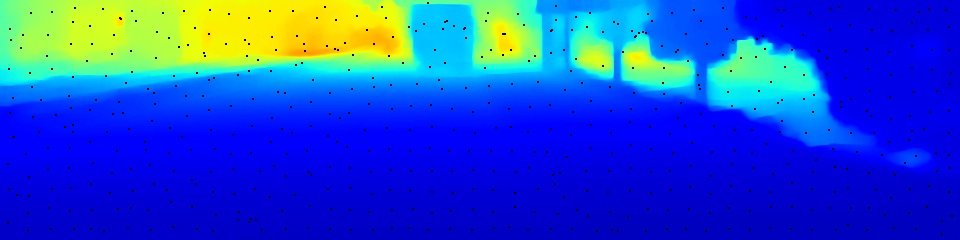}
\put (0,4.2){\color{white}\tiny RMSE: 1559.9}
\put (0,1.2){\color{white}\tiny SPS Sampling + FusionNet Reconstruction}
\end{overpic}
&
\begin{overpic}[width=0.25\linewidth]{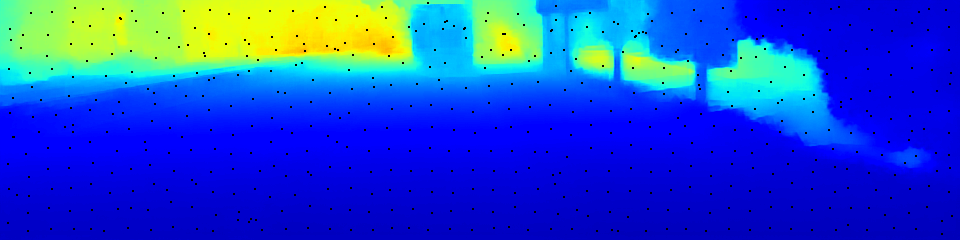}
\put (0,4.2){\color{white}\tiny RMSE: 1902.9}
\put (0,1.2){\color{white}\tiny SPS Sampling + SSNet Reconstruction}
\end{overpic}
&
\begin{overpic}[width=0.25\linewidth]{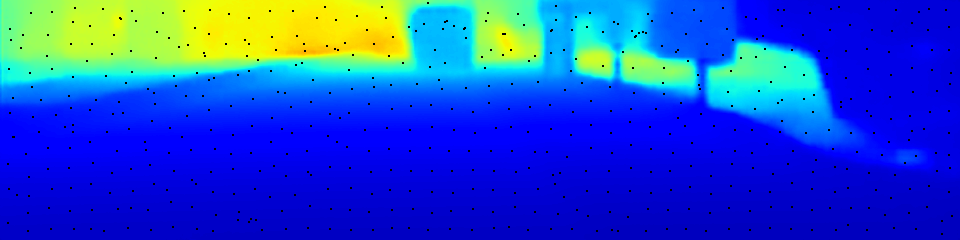}
\put (0,4.2){\color{white}\tiny RMSE: 1848.4}
\put (0,1.2){\color{white}\tiny SPS Sampling + $NetE$ Reconstruction}
\end{overpic}
&
\begin{overpic}[width=0.25\linewidth]{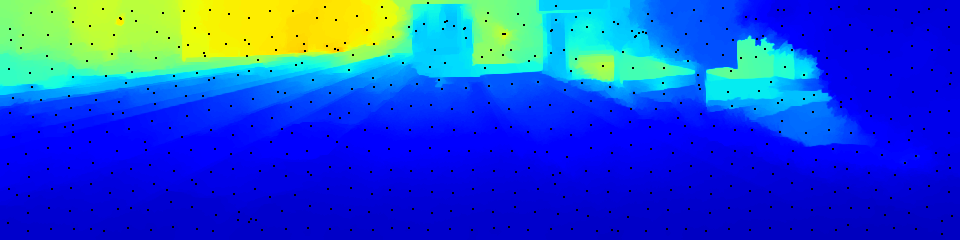}
\put (0,4.2){\color{white}\tiny RMSE: 3280.7}
\put (0,1.2){\color{white}\tiny SPS Sampling + Colorization Reconstruction}
\end{overpic}
\\
\begin{overpic}[width=0.25\linewidth]{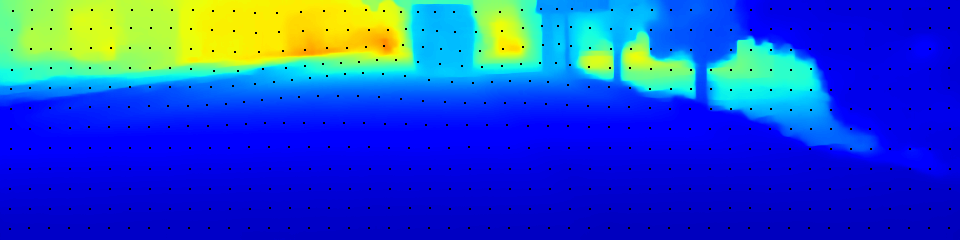}
\put (0,4.2){\color{white}\tiny RMSE: 1561.6}
\put (0,1.2){\color{white}\tiny DAL Sampling + FusionNet Reconstruction}
\end{overpic}
&
\begin{overpic}[width=0.25\linewidth]{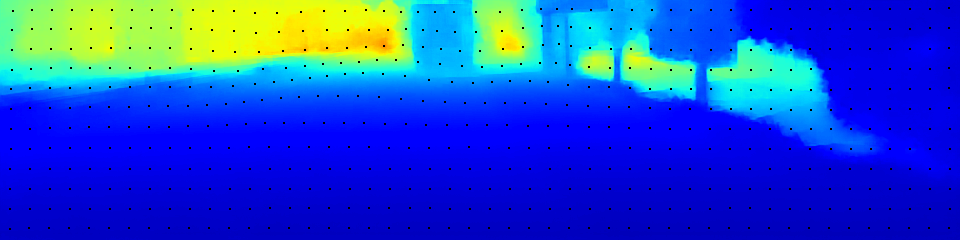}
\put (0,4.2){\color{white}\tiny RMSE: 1883.8}
\put (0,1.2){\color{white}\tiny DAL Sampling + SSNet Reconstruction}
\end{overpic}
&
\begin{overpic}[width=0.25\linewidth]{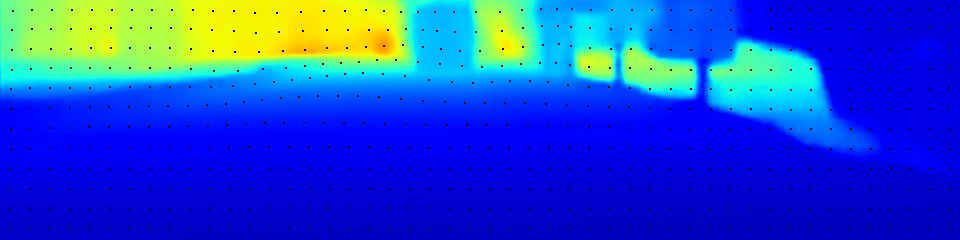}
\put (0,4.2){\color{white}\tiny RMSE: 1839.4}
\put (0,1.2){\color{white}\tiny DAL Sampling + $NetE$ Reconstruction}
\end{overpic}
&
\begin{overpic}[width=0.25\linewidth]{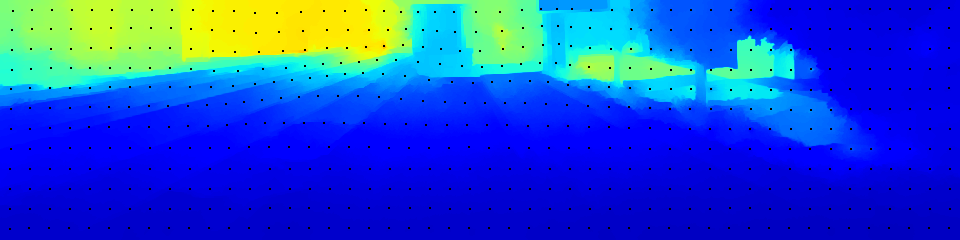}
\put (0,4.2){\color{white}\tiny RMSE: 3376.6}
\put (0,1.2){\color{white}\tiny DAL Sampling + Colorization Reconstruction}
\end{overpic}
\\
\begin{overpic}[width=0.25\linewidth]{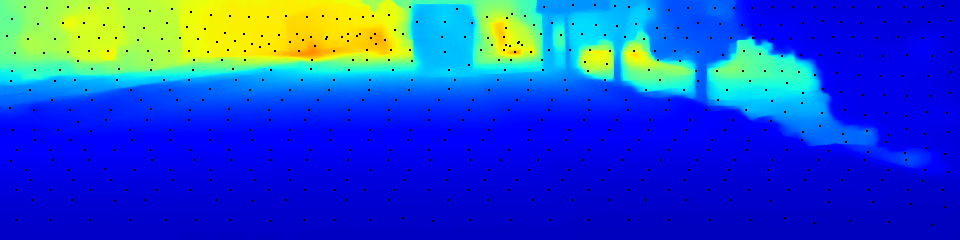}
\put (0,4.2){\color{white}\tiny RMSE: 1248.5}
\put (0,1.2){\color{white}\tiny $NetM$ Sampling + FusionNet Reconstruction}
\end{overpic}
&
\begin{overpic}[width=0.25\linewidth]{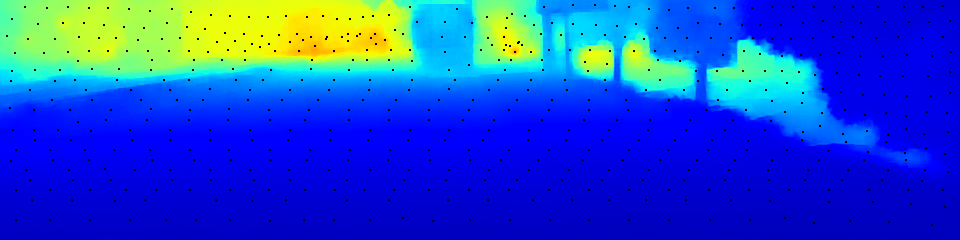}
\put (0,4.2){\color{white}\tiny RMSE: 1531.2}
\put (0,1.2){\color{white}\tiny $NetM$ Sampling + SSNet Reconstruction}
\end{overpic}
&
\begin{overpic}[width=0.25\linewidth]{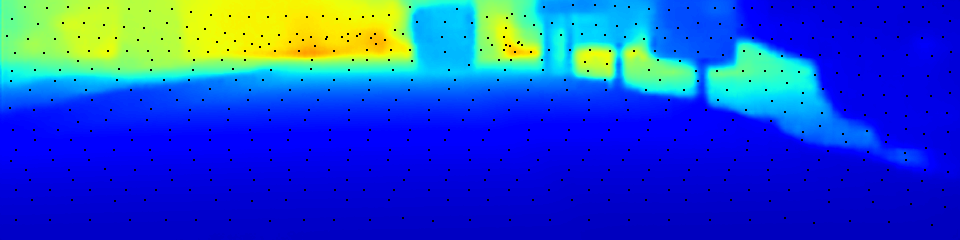}
\put (0,4.2){\color{white}\tiny RMSE: 1505.1}
\put (0,1.2){\color{white}\tiny $NetM$ Sampling + $NetE$ Reconstruction}
\end{overpic}
&
\begin{overpic}[width=0.25\linewidth]{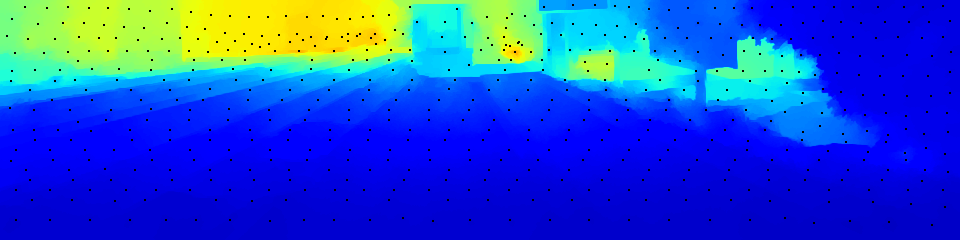}
\put (0,4.2){\color{white}\tiny RMSE: 3028.1}
\put (0,1.2){\color{white}\tiny $NetM$ Sampling + Colorization Reconstruction}
\end{overpic}
\\
\end{tabular}

\end{center}
\caption{Visual comparison of the estimated depth maps. Random, Poisson, SPS, DAL, and $NetM$ sampling masks at sampling rate $c=0.25\%$ are applied and shown in the $2^{nd}-6^{th}$ rows, respectively. The first row includes the RGB image and the ground truth depth map. Sampling locations are indicated using black dots. FusionNet, SSNet, $NetE$ and Colorization depth estimation methods are used to perform depth estimation and generate the depth maps of $1^{st}-4^{th}$ columns, respectively. RMSE is computed for each depth map with respect to the ground truth depth map.}

\label{fig:visual_compare}
\end{figure*}

During the training of $NetE$, we follow Ma \emph{et~al.}'s setup~\cite{mal2018sparse}. The batch size is set equal to $16$. The ResNet encoder is initialized with pretrained weights using the ImageNet dataset~\cite{russakovsky2015imagenet}. Stochastic gradient descent (SGD) optimizer with momentum $0.9$ is used. We train $100$ epochs in total. The learning rate is set to be equal to $0.01$ at first and reduced by $80\%$ at every $25$ epochs. $NetE$ is trained individually under different sampling rates $c=1\%, 0.25\%$ and $0.0625\%$ using random sampling masks. We also train FusionNet~\cite{van2019sparse} and SSNet~\cite{ma2019self} under different sampling rates using random sampling masks. The same training procedure in their original papers are used. They serve as alternative depth estimation methods.

We test the proposed sampling algorithm under $3$ sampling rates, $c=1\%, 0.25\%$ and $0.0625\%$. For the KITTI depth completion dataset, they correspond to $N_s=2304, 576$ and $144$ depth samples (superpixels) in the $240\times 940$ image. For the NYU-Depth-V2 dataset, they correspond to $N_s=768, 192$ and $48$ depth samples (superpixels) in the $240\times 320$ image. $NetM$ is configured to output the desired number of samples. During the training of $NetM$, we pretrain it using the SLIC loss. $m$ in Equation~\ref{e:e3} is set equal to be $1$. ADAM optimizer~\cite{kingma2014adam} is applied. Learning rate is set to be $5\times 10^{-5}$. We train $100$ epochs in total.

After $NetM$ is initialized, we finally jointly train $NetM$ and $NetE$ according to Figure~\ref{fig:proposed_pipeline}. Loss defined in Equation~\ref{e:e8} is optimized with $q$ equal to $10^{-6}$, resulting in $\mathcal{L}_{depth}$ being equal to about $10$ times of $q\cdot\mathcal{L}_{SLIC}$ in value. The window size of the soft depth sampling module is equal to $5$. Temperature $t$ defined in Equation~\ref{e:e6} decreases from $1.0$ to $0.1$ linearly during training. We experimentally find that $NetM$'s performance is not sensitive to the SSA related settings, such as the window size, initial temperature and temperature decay policy. Batch size is set equal to $8$ and this is the largest batch size we can use for both $NetM$ and $NetE$ in an NVIDIA 2080Ti GPU (11GB memory). As discussed in Section~\ref{sec:training_procedures}, $NetE$ is fixed during the training to make $NetM$ generalize well to other depth estimation methods. Learning rate of $NetM$ is assigned to be equal to $10^{-4}$ and is reduced by $50\%$ every 10 epochs. SGD optimizer with momentum $0.9$ is used. We found that 50 epochs in total are adequate for converge.

Our proposed adaptive depth sampling framework is implemented in PyTorch and our implementation is available at: \url{https://github.com/usstdqq/adaptive-depth-sensing}.

\subsection{Performance on Adaptive Depth Sensing and Estimation}
\label{sec:performance_sensing_estimation}

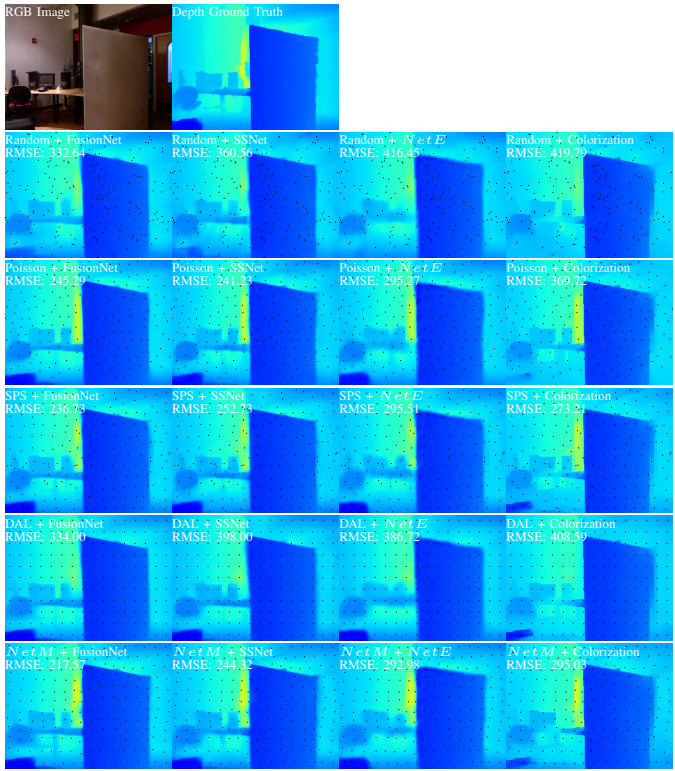
\begin{figure}[!t]
\tiny
\begin{center}
\setlength{\tabcolsep}{0.1pt}
\begin{tabular}{c c c c }

\begin{overpic}[width=0.25\linewidth]{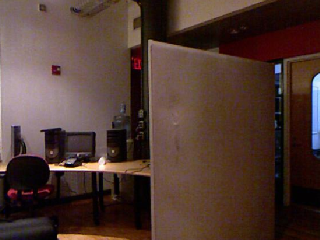}
\put (0,68){\color{white}\tiny RGB Image}
\end{overpic}
&
\begin{overpic}[width=0.25\linewidth]{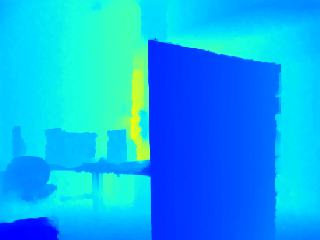}
\put (0,68){\color{white}\tiny Depth Ground Truth}
\end{overpic}
&
& \\
\begin{overpic}[width=0.25\linewidth]{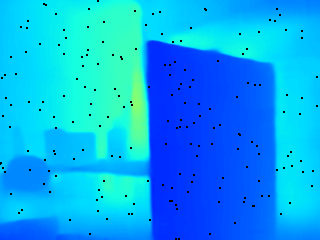}
\put (0,60){\color{white}\tiny RMSE: 332.64}
\put (0,68){\color{white}\tiny Random + FusionNet}
\end{overpic}
&
\begin{overpic}[width=0.25\linewidth]{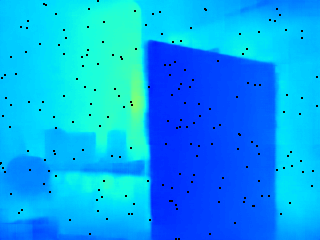}
\put (0,60){\color{white}\tiny RMSE: 360.56}
\put (0,68){\color{white}\tiny Random + SSNet}
\end{overpic}
&
\begin{overpic}[width=0.25\linewidth]{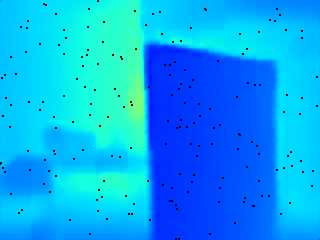}
\put (0,60){\color{white}\tiny RMSE: 416.45}
\put (0,68){\color{white}\tiny Random + $NetE$}
\end{overpic}
&
\begin{overpic}[width=0.25\linewidth]{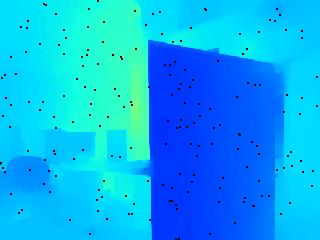}
\put (0,60){\color{white}\tiny RMSE: 419.79}
\put (0,68){\color{white}\tiny Random + Colorization}
\end{overpic}
\\
\begin{overpic}[width=0.25\linewidth]{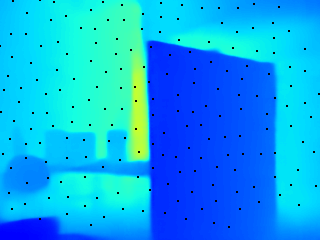}
\put (0,60){\color{white}\tiny RMSE: 245.29}
\put (0,68){\color{white}\tiny Poisson + FusionNet}
\end{overpic}
&
\begin{overpic}[width=0.25\linewidth]{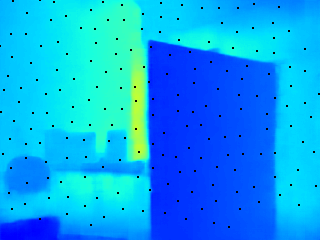}
\put (0,60){\color{white}\tiny RMSE: 241.23}
\put (0,68){\color{white}\tiny Poisson + SSNet}
\end{overpic}
&
\begin{overpic}[width=0.25\linewidth]{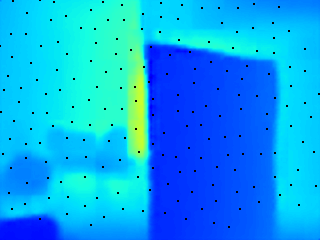}
\put (0,60){\color{white}\tiny RMSE: 295.27}
\put (0,68){\color{white}\tiny Poisson + $NetE$}
\end{overpic}
&
\begin{overpic}[width=0.25\linewidth]{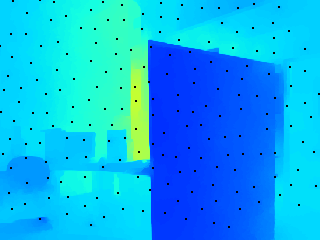}
\put (0,60){\color{white}\tiny RMSE: 369.72}
\put (0,68){\color{white}\tiny Poisson + Colorization}
\end{overpic}
\\
\begin{overpic}[width=0.25\linewidth]{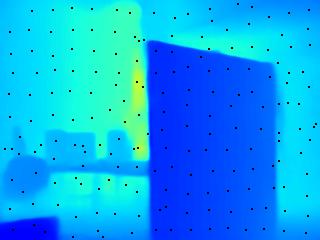}
\put (0,60){\color{white}\tiny RMSE: 236.73}
\put (0,68){\color{white}\tiny SPS + FusionNet}
\end{overpic}
&
\begin{overpic}[width=0.25\linewidth]{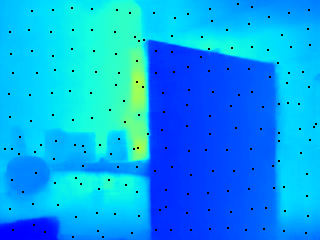}
\put (0,60){\color{white}\tiny RMSE: 252.73}
\put (0,68){\color{white}\tiny SPS + SSNet}
\end{overpic}
&
\begin{overpic}[width=0.25\linewidth]{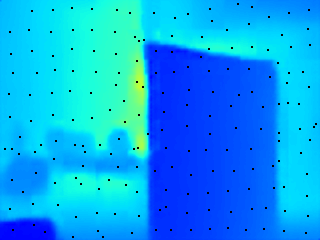}
\put (0,60){\color{white}\tiny RMSE: 295.51}
\put (0,68){\color{white}\tiny SPS + $NetE$}
\end{overpic}
&
\begin{overpic}[width=0.25\linewidth]{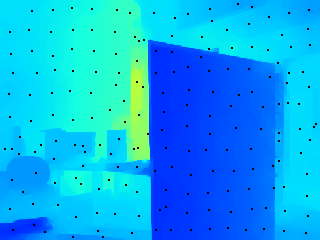}
\put (0,60){\color{white}\tiny RMSE: 273.21}
\put (0,68){\color{white}\tiny SPS + Colorization}
\end{overpic}
\\
\begin{overpic}[width=0.25\linewidth]{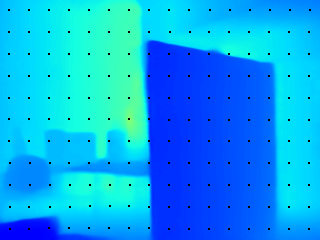}
\put (0,60){\color{white}\tiny RMSE: 334.00}
\put (0,68){\color{white}\tiny DAL + FusionNet}
\end{overpic}
&
\begin{overpic}[width=0.25\linewidth]{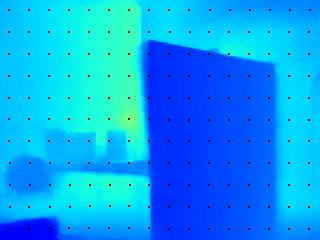}
\put (0,60){\color{white}\tiny RMSE: 398.00}
\put (0,68){\color{white}\tiny DAL + SSNet}
\end{overpic}
&
\begin{overpic}[width=0.25\linewidth]{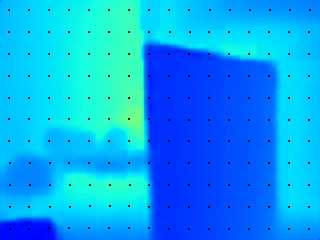}
\put (0,60){\color{white}\tiny RMSE: 386.72}
\put (0,68){\color{white}\tiny DAL + $NetE$}
\end{overpic}
&
\begin{overpic}[width=0.25\linewidth]{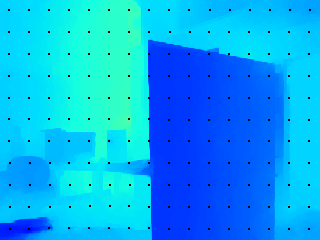}
\put (0,60){\color{white}\tiny RMSE: 408.59}
\put (0,68){\color{white}\tiny DAL + Colorization}
\end{overpic}
\\
\begin{overpic}[width=0.25\linewidth]{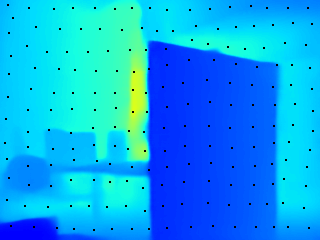}
\put (0,60){\color{white}\tiny RMSE: 217.57}
\put (0,68){\color{white}\tiny $NetM$ + FusionNet}
\end{overpic}
&
\begin{overpic}[width=0.25\linewidth]{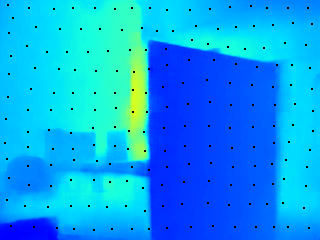}
\put (0,60){\color{white}\tiny RMSE: 244.32}
\put (0,68){\color{white}\tiny $NetM$ + SSNet}
\end{overpic}
&
\begin{overpic}[width=0.25\linewidth]{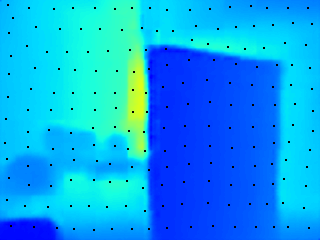}
\put (0,60){\color{white}\tiny RMSE: 292.98}
\put (0,68){\color{white}\tiny $NetM$ + $NetE$}
\end{overpic}
&
\begin{overpic}[width=0.25\linewidth]{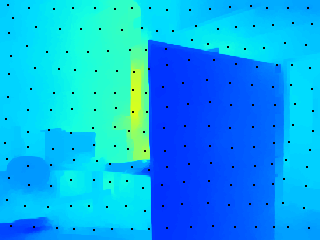}
\put (0,60){\color{white}\tiny RMSE: 295.03}
\put (0,68){\color{white}\tiny $NetM$ + Colorization}
\end{overpic}
\\
\end{tabular}

\end{center}
\caption{Visual comparison of the estimated depth maps. Random, Poisson, SPS, DAL, and $NetM$ sampling methods at sampling rate $c=0.25\%$ are applied and shown in the $2^{nd}-6^{th}$ rows, respectively. The first row includes the RGB image and the ground truth depth map. Sampling locations are indicated using black dots. FusionNet, SSNet, $NetE$ and Colorization depth estimation methods are used to perform depth estimation and generate the depth maps of $1^{st}-4^{th}$ columns, respectively. RMSE is computed for each depth map with respect to the ground truth depth map.}

\label{fig:visual_compare_nyuv2}
\end{figure}

For the adaptive depth sampling and estimation task, we demonstrate the advantages of our proposed adaptive sampling mask $NetM$, over the use of random, uniform grid and Poisson~\cite{bridson2007fast} sampling masks, as well as other state-of-the-art adaptive depth sampling methods, such as SuperPixel Sampler (SPS)~\cite{wolff2020super} and Deep Adaptive Lidar (DAL)~\cite{bergman2020deep}. 

$NetM$ is initialized by RGB images according to FCN~\cite{yang2020superpixel}. To show the effectiveness of the proposed $NetE$ and $NetM$ joint training method, we also compare with the sampling mask computed by the initialized $NetM$. The sampling method is denoted as FCN.

To illustrate $NetM$ can generalize across datasets, we train $NetM$ on the NYU-Depth-V2 dataset and test it on the KITTI dataset. The depth sampling method is noted as $NetM-NYU$ when testing on the KITTI dataset. Similarly, we train $NetM$ on the KITTI dataset and test it on the NYU-Depth-V2 dataset. The sampling method is noted as $NetM-KITTI$ when testing on the NYU-Depth-V2 dataset. Noted that $NetM$ is fully convolutional, so it is straightforward to test on different image resolution.

Random, Uniform Grid, Poisson, SPS~\cite{wolff2020super}, DAL~\cite{bergman2020deep}, FCN~\cite{yang2020superpixel} and proposed $NetM$ (including $NetM-KITTI$ and $NetM-NYU$) sampling methods are applied to the test images. Sampling rates $c=1\%, 0.25\%$ and $0.0625\%$ are tested. For the depth estimation methods, DL-based methods $NetE$~\cite{mal2018sparse}, FusionNet~\cite{van2019sparse}, SSNet~\cite{ma2019self} and traditional method Colorization~\cite{levin2004colorization} are used to estimate the fully sampled depth map from the sampled depth map and RGB image. It's noted that all the DL-based depth estimation methods are trained using random sampling masks and the same training set of either KITTI depth completion or NYU-Depth-V2 dataset.

For the KITTI depth completion dataset, the average Root Mean Square Error (RMSE) and Mean Absolute Error (MAE) over all $3426$ test frames are shown in Table~\ref{table:depth_sampling_estimation_table}. First, under all three sampling rates, the proposed $NetM$ mask outperforms the random, Poisson, Uniform Grid, SPS, DAL and FCN masks consistently over all depth estimation methods in terms of RMSE and MAE. This demonstrates the effectiveness of our proposed adaptive depth sampling network. Furthermore, $NetM$ is jointly trained with $NetE$ and it still performs well with other depth estimation methods, demonstrating that it can generalize well to other depth estimation methods. The performance advantage of $NetM$ is not tied to any specific depth estimation method. Finally, it can be concluded that the smaller the sampling rate, the larger the advantage of $NetM$ compared to other sampling algorithms. This implies that $NetM$ is able to handle challenging depth sampling tasks (extremely low sampling rates).

The depth sampling and reconstruction performance on the NYU-Depth-V2 dataset are shown in Table~\ref{table:depth_sampling_estimation_table_nyu}. Similar conclusions as the ones from the KITTI depth completion dataset can be drawn. It is noticed that SPS method is comparable to $NetM$. NYU-Depth-V2 is an indoor dataset with a maximum $10$m depth range. The variance of the depth map is small compared to the KITTI depth completion dataset, so the even spatial distribution prior in SPS works well here. Moreover, NYU-Depth-V2 dataset is captured by Microsoft Kinect. The ground truth depth maps have low spatial resolution ($240\times320$) and are relative noisy. Thus some improvement on fine details in the estimated depth map is not reflected. Nevertheless, $NetM$ outperforms SPS when sampling rate is low and more advanced DL-based depth completion algorithms are applied.

According to Table~\ref{table:depth_sampling_estimation_table} and Table~\ref{table:depth_sampling_estimation_table_nyu}, FusionNet~\cite{van2019sparse} has the best depth estimation accuracy under various of sampling masks. FusionNet extracts both global and local information and is more complex than $NetE$. Depth estimation from RGB image and sparse depth input is an active research topic. $NetM$ is able to generalize with other depth estimation methods than $NetE$, so it is able to benefit from the latest depth estimation algorithms. 

The visual quality comparison of various sampling strategies and depth estimation methods is shown in Figure~\ref{fig:visual_compare} and Figure~\ref{fig:visual_compare_nyuv2}. For sampling rate equal to $c=0.25\%$, we can observe the advantages of the proposed $NetM$ mask over all other sampling masks by comparing the resulting depth maps by the same depth estimation algorithm. In Figure~\ref{fig:visual_compare}, $NetM$ samples densely around the end of the road, trees and billboard, resulting in accurate depth estimation in such areas. Compared to other adaptive sampling algorithms, such as SPS and DAL, $NetM$ samples more densely on distance objects, making the estimated depth more accurate. SPS uses SLIC~\cite{achanta2012slic} to segment the RGB image and such segmentation can not obtain distance information from the RGB images. DAL estimates a smooth motion field to warp a regular sampling grid. When the scene is complicated, it is not flexible enough to warp a regular sampling grid to optimal location. In Figure~\ref{fig:visual_compare_nyuv2} , $NetM$ samples the distant vertical structure, as well as the table and chair on the left side. More detailed depth map reconstruction in those areas is obtained. Compared to SPS, $NetM$ tends to sample uniformly on the wall regions, while still capture the shape of the foreground objects.

\begin{figure}[h]
\scriptsize
\begin{center}
\begin{tabular}{r l }
\rotatebox{90}{\ $NetM-NYU$} &
\includegraphics[width=0.9\linewidth]{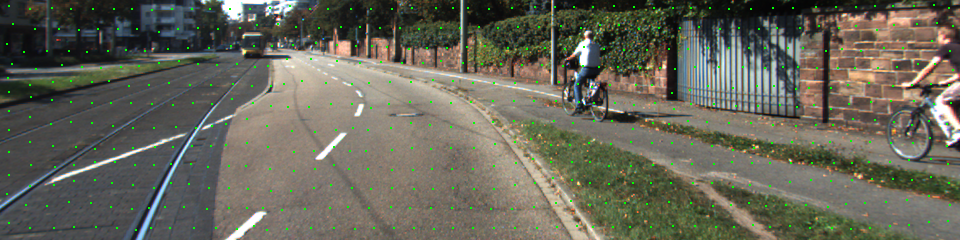}\\
\rotatebox{90}{\ \ \ \ \ \ \ $NetM$} &
\includegraphics[width=0.9\linewidth]{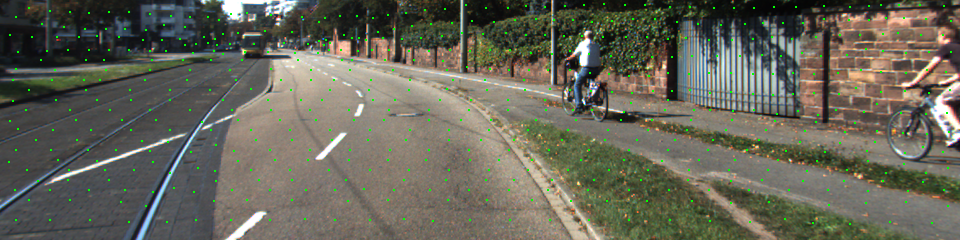}\\
\end{tabular}
\end{center}
\caption{Visual comparison of $NetM-NYU$ and $NetM$'s sampling location on the KITTI dataset. Sampling locations are plotted in green.}
\label{fig:cross_test_rgb_mask_visual_compare_kitti}
\end{figure}

\begin{figure}[h]
\scriptsize
\begin{center}
\begin{tabular}{c c }
\includegraphics[width=0.45\linewidth]{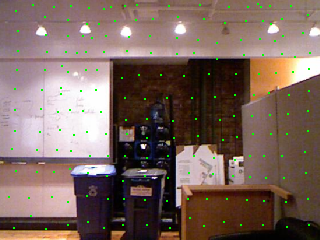} &
\includegraphics[width=0.45\linewidth]{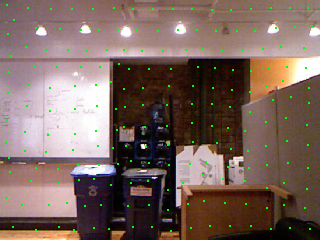}\\
$NetM-KITTI$ & $NetM$
\end{tabular}
\end{center}
\caption{Visual comparison of $NetM-KITTI$ and $NetM$'s sampling location on the NYU-Depth-V2 dataset. Sampling locations are plotted in green.}
\label{fig:cross_test_rgb_mask_visual_compare_nyu}
\end{figure}

According to Table~\ref{table:depth_sampling_estimation_table}, $NetM-NYU$ is the second best performer in the KITTI dataset. From Table~\ref{table:depth_sampling_estimation_table_nyu}, in the NYU-Depth-V2 dataset, $NetM-KITTI$ outperforms all the other sampling methods except SPS and $NetM$. This demonstrates that the proposed $NetM$ is able to generalize across different datasets. We visualize the sampling location difference between $NetM-NYU$ and $NetM$ in the KITTI dataset in Figure~\ref{fig:cross_test_rgb_mask_visual_compare_kitti}. It can be found that $NetM$ samples densely on the upper part of the image compared to $NetM-NYU$. $NetM-NYU$ does not learn the prior knowledge that upper part of the image has larger depth values and should be sampled more densely from the indoor NYU-Depth-V2 dataset. However, $NetM-NYU$ is still able to sample on such objects as the bus, cyclist and poles. The sampling location difference between $NetM-KITTI$ and $NetM$ on the NYU-Depth-V2 dataset is shown in Figure~\ref{fig:cross_test_rgb_mask_visual_compare_nyu}. $NetM-KITTI$ learns the prior knowledge that upper part of the image is more important from the KITTI dataset and samples densely on the upper part of the image, resulting sub-optimal reconstruction accuracy compared to $NetM$.

\begin{figure}[h]
\scriptsize
\begin{center}
\begin{tabular}{r l }
\rotatebox{90}{\ \ \ \ \ \ \ SLIC} &
\includegraphics[width=0.9\linewidth]{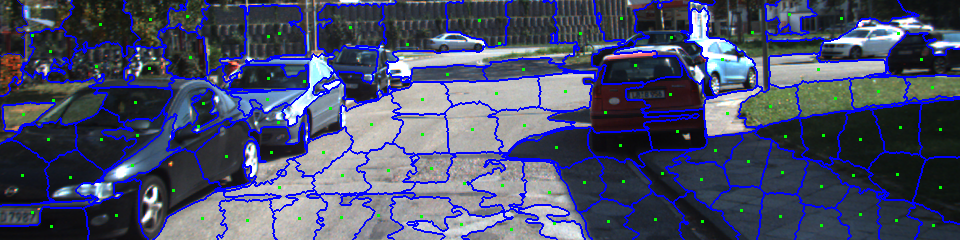}\\
\rotatebox{90}{\ \ \ \ \ \ \ FCN} &
\includegraphics[width=0.9\linewidth]{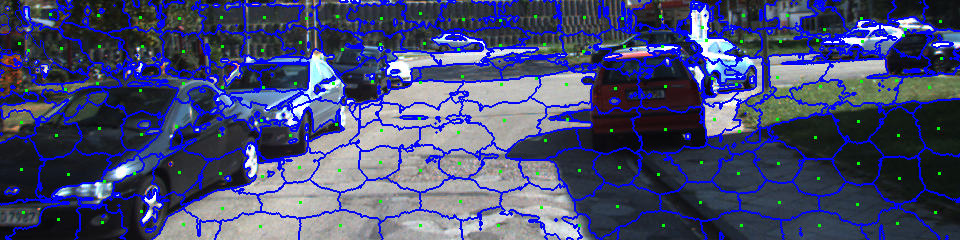}\\
\rotatebox{90}{\ \ \ \ \ \ \ $NetM$} &
\includegraphics[width=0.9\linewidth]{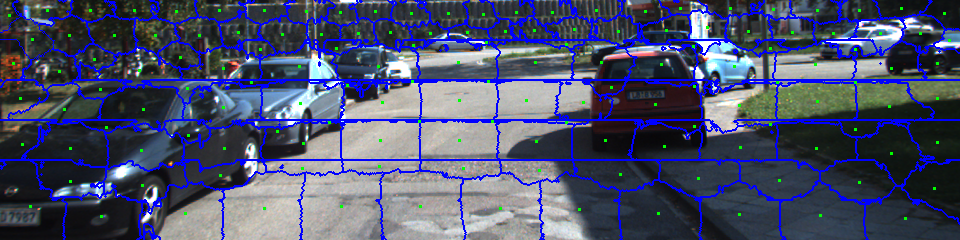}\\
\end{tabular}
\end{center}
\caption{Visual comparison of different superpixel segmentation and sampling location. Segmentation boundaries are plotted in blue and sampling locations are plotted in green.}
\label{fig:super_pixel_compare}
\end{figure}

$NetM$ is initialized with RGB images trained FCN~\cite{yang2020superpixel} superpixel network using the SLIC loss (Equation~\ref{e:e3}). SPS~\cite{wolff2020super} uses the SLIC superpixel technique to segment the RGB images. Sampling locations are determined by the weighted mass center of superpixels. Different superpixel segmentations result in different sampling quality. In Figure~\ref{fig:super_pixel_compare}, we visualize the superpixel segmentation results and the derived sampling locations for SLIC, FCN and $NetM$ when $c=0.0625\%$. SLIC and FCN segment the input RGB image based on the color similarity and preserve spatial compactness. The segmentation density is spatially homogeneous. $NetM$ is jointly trained with $NetE$, thus it has knowledge of distance given the RGB input image. Distance objects in the image are sampled denser. It also segments sparsely the pavement and grass areas. Such near objects as cars are segmented denser compared to the pavement and grass areas. We also observe that $NetM$ segmentation does not preserve color pixel boundaries as well as FCN, which is expected as $NetM$ also minimizes the depth estimation loss besides the SLIC loss in Equation~\ref{e:e8}. According to FCN and $NetM$'s reconstruction accuracy in Table~\ref{table:depth_sampling_estimation_table} and Table~\ref{table:depth_sampling_estimation_table_nyu}, it can be found that $NetM$ always outperforms FCN. This demonstrates the effectiveness of the proposed $NetM$ training mechanism (Equation~\ref{e:e8}).

\subsection{Effectiveness of Soft Sampling Approximation}

\begin{table}[!t]
\begin{center}
\scriptsize
\begin{tabular}{c | c | c | c | c | c | c | c  }\hline\hline
\multicolumn{2}{c|}{}& \multicolumn{2}{c|}{FusionNet} &  \multicolumn{2}{c|}{SSNet} & \multicolumn{2}{c}{$NetE$}\\ \hline
$c$& Kernel & MAE & RMSE & MAE & RMSE & MAE & RMSE  \\ \hline
\multirow{2}{*}{\scriptsize{$1\%$}} & \scriptsize{Bilinear}    & 290.5 & 948.4  & 436.3  & 1086.7 & 383.0 & 1138.8 \\  \cline{2-8}
                                    & \scriptsize{SSA}         & 285.0 & 939.4  & 423.1  & 1074.9 & 380.1 & 1131.3 \\ \hline
\multirow{2}{*}{\scriptsize{$0.25\%$}}&\scriptsize{Bilinear}   & 431.1 & 1285.3 & 590.1  & 1487.5 & 494.1 & 1466.0 \\ \cline{2-8}
                                      &\scriptsize{SSA}        & 404.3 & 1239.7 & 562.2  & 1436.8 & 477.5 & 1422.4  \\ \hline
\multirow{2}{*}{\scriptsize{$0.0625\%$}}&\scriptsize{Bilinear} & 809.1 & 2161.4 & 989.2  & 2457.4 & 781.6 & 2253.0 \\  \cline{2-8}
                                        &\scriptsize{SSA}      & 634.9 & 1732.4 & 778.0  & 1930.5 & 652.2 & 1896.7 \\ \hline \hline
\end{tabular}
\end{center}
\caption{Using SSA and Bilinear kernel during training results different sampling quality of $NetM$.}
\label{table:soft_sampling_compare}
\end{table}

In Section~\ref{sec:soft_sampling_approximation}, we propose the use of SSA to make the sampling process differentiable during training. Such differentiable sampling approximation is necessary to jointly train $NetM$ with $NetE$. Compared to the $2\times 2$ bilinear kernel based differentable sampling in~\cite{bergman2020deep}, the proposed SSA provides better sampling performance. In order to show the advantages of SSA, we replace the SSA sampling of $NetM$ by the bilinear kernel based sampling and perform the exact same training procedures. As demonstrated in Table~\ref{table:soft_sampling_compare}, the lower the sampling rate, the bigger the advantage of SSA over the bilinear kernel sampling. When sampling points are sparse, the gradients derived from a $2\times 2$ local window are too small to train $NetM$ effectively. We empirically found that the $5\times 5$ window size for SSA provides reasonable sampling performance under all sampling rates.

\begin{table}[!b]
\begin{center}
\scriptsize
\begin{tabular}{c | c | c | c | c }\hline\hline
$c$                                      & Sampling & Reconstruction & MAE    & RMSE    \\  \hline
\multirow{4}{*}{\scriptsize{$1\%$}}      & SPS      & SPS            & 406.3  & 1264.2  \\  \cline{2-5} 
                                         & DAL      & DAL            & 550.3  & 1566.7     \\  \cline{2-5} 
                                         & $NetM$   & FusionNet      & 285.0  & 939.4   \\  \cline{2-5}
                                         & $NetM*$  & FusionNet*     & \textbf{284.6}  & \textbf{932.6}   \\  \hline            
\multirow{4}{*}{\scriptsize{$0.25\%$}}   & SPS      & SPS            & 812.7  & 2192.3  \\  \cline{2-5} 
                                         & DAL      & DAL            & 597.7  & 1667.8     \\  \cline{2-5} 
                                         & $NetM$   & FusionNet      & 404.3  & 1239.7  \\  \cline{2-5}
                                         & $NetM*$  & FusionNet*     & \textbf{402.9}  & \textbf{1229.4}  \\  \hline    
\multirow{4}{*}{\scriptsize{$0.0625\%$}} & SPS      & SPS            & 1668.6 & 3891.9  \\  \cline{2-5} 
                                         & DAL      & DAL            & 789.1  & 2104.0     \\  \cline{2-5} 
                                         & $NetM$   & FusionNet      & 634.9  & 1732.3   \\  \cline{2-5}
                                         & $NetM*$  & FusionNet*     & \textbf{631.5}  & \textbf{1721.1}       \\  \hline \hline
\end{tabular}
\end{center}
\caption{End to end depth estimation results comparison.}
\label{table:end_to_end_compare}
\end{table}

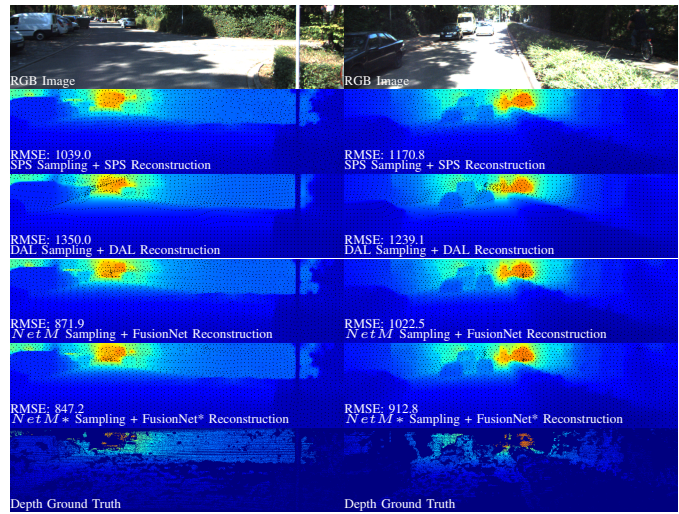
\begin{figure}[h]
\tiny
\begin{center}
\setlength{\tabcolsep}{0.2pt}
\begin{tabular}{c c }

\begin{overpic}[width=0.5\linewidth]{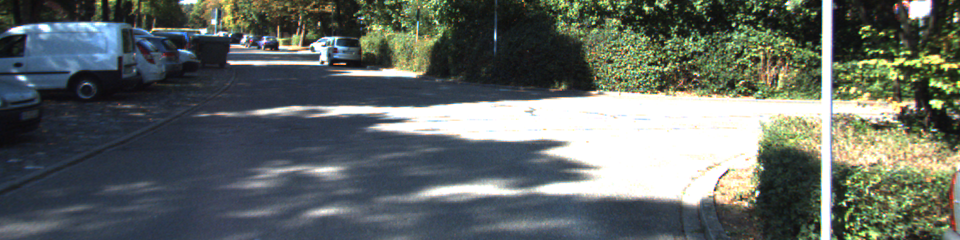}
\put (0,1.2){\color{white}\tiny RGB Image}
\end{overpic}
&
\begin{overpic}[width=0.5\linewidth]{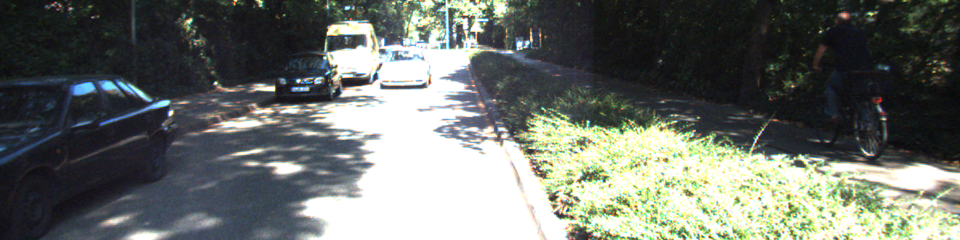}
\put (0,1.2){\color{white}\tiny RGB Image}
\end{overpic}\\

\begin{overpic}[width=0.5\linewidth]{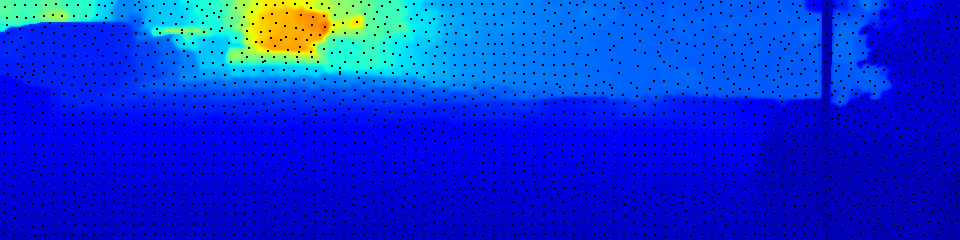}
\put (0,4.2){\color{white}\tiny RMSE: 1039.0}
\put (0,1.2){\color{white}\tiny SPS Sampling + SPS Reconstruction}
\end{overpic}
&
\begin{overpic}[width=0.5\linewidth]{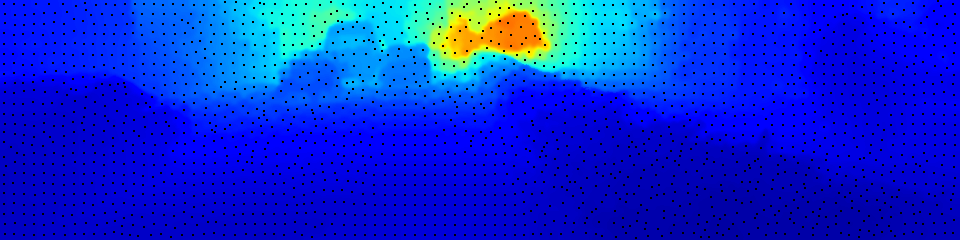}
\put (0,4.2){\color{white}\tiny RMSE: 1170.8}
\put (0,1.2){\color{white}\tiny SPS Sampling + SPS Reconstruction}
\end{overpic}\\

\begin{overpic}[width=0.5\linewidth]{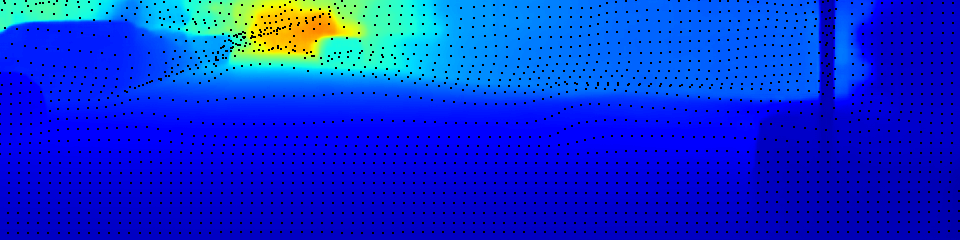}
\put (0,4.2){\color{white}\tiny RMSE: 1350.0}
\put (0,1.2){\color{white}\tiny DAL Sampling + DAL Reconstruction}
\end{overpic}
&
\begin{overpic}[width=0.5\linewidth]{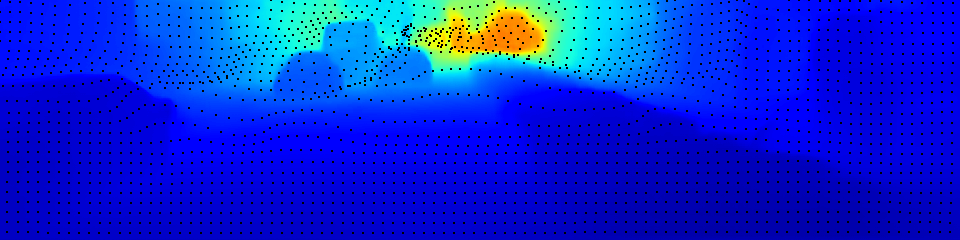}
\put (0,4.2){\color{white}\tiny RMSE: 1239.1}
\put (0,1.2){\color{white}\tiny DAL Sampling + DAL Reconstruction}
\end{overpic}\\

\begin{overpic}[width=0.5\linewidth]{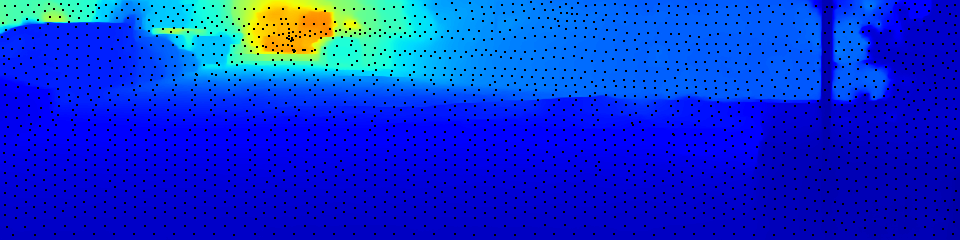}
\put (0,4.2){\color{white}\tiny RMSE: 871.9}
\put (0,1.2){\color{white}\tiny $NetM$ Sampling + FusionNet Reconstruction}
\end{overpic}
&
\begin{overpic}[width=0.5\linewidth]{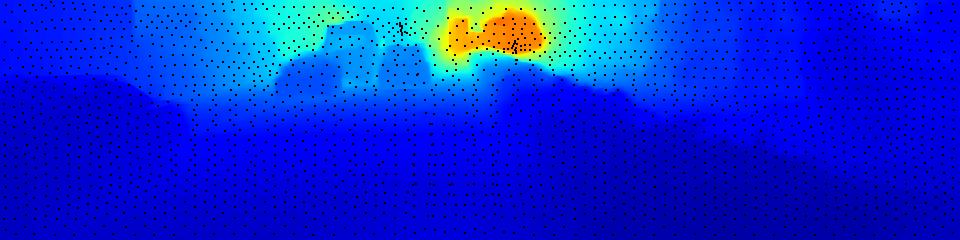}
\put (0,4.2){\color{white}\tiny RMSE: 1022.5}
\put (0,1.2){\color{white}\tiny $NetM$ Sampling + FusionNet Reconstruction}
\end{overpic}\\

\begin{overpic}[width=0.5\linewidth]{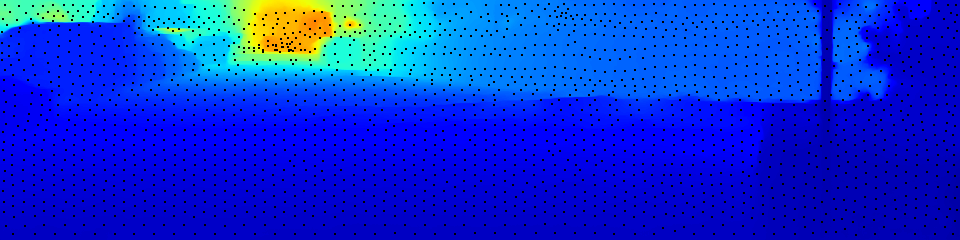}
\put (0,4.2){\color{white}\tiny RMSE: 847.2}
\put (0,1.2){\color{white}\tiny $NetM*$ Sampling + FusionNet* Reconstruction}
\end{overpic}
&
\begin{overpic}[width=0.5\linewidth]{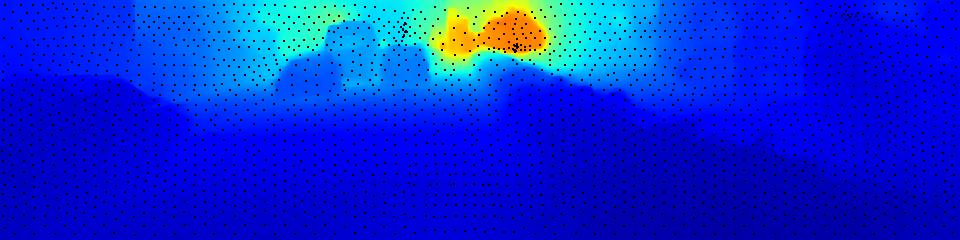}
\put (0,4.2){\color{white}\tiny RMSE: 912.8}
\put (0,1.2){\color{white}\tiny $NetM*$ Sampling + FusionNet* Reconstruction}
\end{overpic}\\

\begin{overpic}[width=0.5\linewidth]{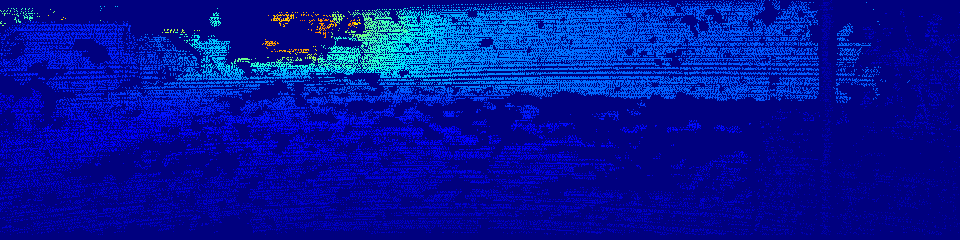}
\put (0,1.2){\color{white}\tiny Depth Ground Truth}
\end{overpic}
&
\begin{overpic}[width=0.5\linewidth]{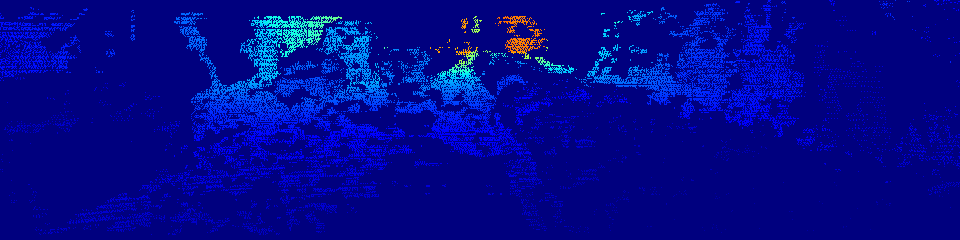}
\put (0,1.2){\color{white}\tiny Depth Ground Truth}
\end{overpic}\\

\end{tabular}

\end{center}
\caption{Visual comparison of different depth sampling and estimation methods.}

\label{fig:visual_compare_end_to_end}
\end{figure}

\subsection{End To End Depth Estimation Performance}

\begin{table*}[h]
\scriptsize
\centering
\begin{tabular}{c | c | c | c | c | c | c | c | c | c | c } \hline \hline
        & \multicolumn{10}{c}{KITTI Depth Completion} \\  \cline{2-11}
        & & \multicolumn{3}{c|}{c=1\%} & \multicolumn{3}{c|}{c=0.25\%} & \multicolumn{3}{c}{c=0.0625\%} \\  \cline{2-11}
        & device & time (ms)   & FLOPs & \#params  & time (ms) & FLOPs & \#params & time (ms) & FLOPs & \#params  \\ \hline
Poisson & CPU   & 68.6  & -      & -     & 18.0  & -      & -     & 5.1   & -      & -  \\ \hline
SPS     & CPU   & 399.8 & -      & -     & 301.0 & -      & -     & 244.5 & -      & -  \\ \hline
DAL     &CPU    & 460.9  & 149.8G & 42.4M & 439.3  & 149.8G & 42.4M & 464.8  & 149.8G & 42.4M \\ \hline
DAL     &GPU    & 97.9  & 149.8G & 42.4M & 44.5  & 149.8G & 42.4M & 34.2  & 149.8G & 42.4M \\ \hline
$NetM$  &CPU  & 143.8  & 32.0G  & 2.3M  & 143.7  & 32.0G  & 2.3M  & 147.5  & 32.0G  & 2.3M \\ \hline
$NetM$  &GPU  & 20.1  & 32.0G  & 2.3M  & 21.5  & 32.0G  & 2.3M  & 23.4  & 32.0G  & 2.3M \\ \hline\hline
        & \multicolumn{10}{c}{NYU-Depth-V2} \\  \cline{2-11}
        & & \multicolumn{3}{c|}{c=1\%} & \multicolumn{3}{c|}{c=0.25\%} & \multicolumn{3}{c}{c=0.0625\%} \\  \cline{2-11}
        & device & time (ms) & FLOPs & \#params  & time (ms) & FLOPs & \#params & time (ms) & FLOPs & \#params  \\ \hline
Poisson &CPU & 22.3  & -     & -     & 6.4  & -     & -     & 2.0  & -     & -  \\ \hline
SPS     &CPU     & 126.1 & -     & -     & 88.7 & -     & -     & 74.6 & -     & -  \\ \hline
DAL     &CPU     & 159.8  & 50.0G & 42.4M & 158.0 & 50.0G & 42.4M & 164.7 & 50.0G & 42.4M \\ \hline
DAL     &GPU     & 42.8  & 50.0G & 42.4M & 24.9 & 50.0G & 42.4M & 21.3 & 50.0G & 42.4M \\ \hline
$NetM$  &CPU  & 46.0   & 5.7G  & 2.3M  & 48.5 & 5.7G  & 2.3M  & 45.6 & 5.7G  & 2.3M \\ \hline
$NetM$  &GPU  & 9.9   & 5.7G  & 2.3M  & 10.6 & 5.7G  & 2.3M  & 10.1 & 5.7G  & 2.3M \\ \hline\hline
\end{tabular}
\smallskip
\caption{Computation time, FLOPs and model size comparison between the Poisson, SPS, DAL and $NetM$ sampling methods. Notice that Poisson and SPS are tested on CPU, while DL-based methods DAL and $NetM$ are tested on both CPU and GPU.}
\label{table:run_speed_compare}
\end{table*}

In Table~\ref{table:depth_sampling_estimation_table}, FusionNet~\cite{van2019sparse} achieves the best depth estimation performance under various of sampling masks. The proposed global and location information fusion is effective and the network size is considerably larger than $NetE$~\cite{mal2018sparse}. Best depth sampling and estimation results are obtained using $NetM$ sampling and FusionNet depth estimation under all sampling rates. It is noted that $NetM$ is trained jointly with $NetE$ and FusionNet is trained using random masks. Similar to DAL, $NetM$ and FusionNet can also be optimized simultaneously. Starting from the $NetE$ trained $NetM$ and random mask trained FusionNet, we alternatively train $NetM$ and FusionNet and denote the trained networks by $NetM*$ and FusionNet*, respectively. The joint depth sampling and reconstruction results are shown in Table~\ref{table:end_to_end_compare}. We also compare with the sampling and reconstruction methods proposed in SPS and DAL. $NetM*$ with FusionNet* slightly outperforms $NetM$ with FusionNet and achieves the best accuracy. Utilizing random sampling masks during the training of depth estimation methods (FusionNet, SSNet, $NetE$) makes the methods robust to other sampling patterns in testing. We also found that $NetM$ trained using different depth estimation methods has similar sampling patterns. So simultaneously training the sampling and reconstruction networks improves the results slightly.

In Figure~\ref{fig:visual_compare_end_to_end}, we visually compare the end to end depth sampling and reconstruction results. In the $2$ test scenes, $NetM*$ with FusionNet* properly sample and reconstruct distant and thin objects, resulting in the best accuracy compared to other methods. With the developing depth estimation algorithms, we can integrate better depth estimation methods into our system. We show in Section~\ref{sec:performance_sensing_estimation} that the performance advantages of $NetM$ can generalize well to other than $NetE$ depth estimation methods.

\subsection{Running Speed}
Apart from the superior adaptive depth sampling quality to other state-of-the-art methods, the proposed sampling method also has advantages in fast computation efficiency. In this section we evaluate the computation efficiency of different methods. The test images from both the KITTI depth completion dataset ($240 \times 960$) and the NYU-Depth-V2 dataset ($240 \times 360$) are used. Non DL-based methods, Poisson and SPS, are tested using one Intel i9-9820X CPU with 64GB memory. DL-based methods, DAL and $NetM$, are tested using the same CPU as well as one NVIDIA 2080Ti GPU with 11GB memory. We also measure the floating point operations (FLOPs) and number of parameters of both models. All methods are tested on the same test image set under 3 different sampling rates for 100 times and the average run time in milliseconds (ms) are reported in Table~\ref{table:run_speed_compare}. It's noticed that comparing to DAL, $NetM$ has smaller model size and faster run time. Also $NetM$'s run time is almost constant under different sampling rates, different from the other three methods. Such fast computation efficiency property makes $NetM$ practical for a real time adaptive depth sensing system. $NetM$ is also faster than Poisson, SPS and DAL when running on the same CPU.

\subsection{Temporal Registration Issue}
All the experiments above assume the RGB images and the sampled depth maps are captured at the exact same time instant. Due to the system response time, $NetM$ computation time, etc, there are temporal registration issue between the sampled depth maps and the RGB images in practice. To make the adaptive depth sampling method practical, it is important to understand how $NetM$'s sampling performance degrades with respect to the capture time delay $\Delta t$ (between the sampled depth map and RGB image). Noted that the Random, Uniform Grid and Poisson sampling masks are free from such temporal registration issue, because the sampling masks are independent from the RGB images. In this section, we performed 2 sets of experiments to simulate the temporal registration issues.

\begin{figure}[h]
\scriptsize
\begin{center}
\begin{tabular}{c c }
\includegraphics[width=0.475\linewidth]{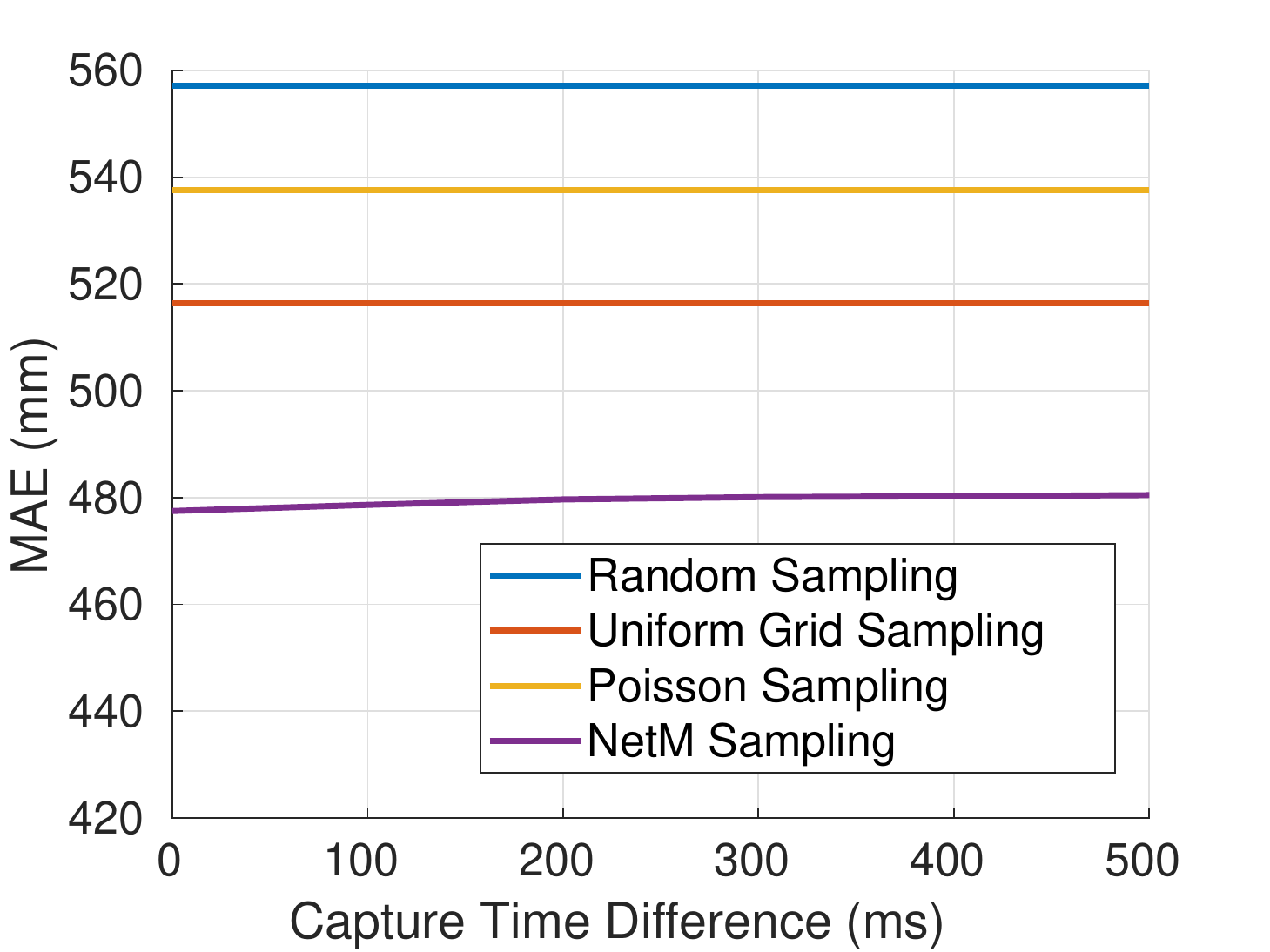} &
\includegraphics[width=0.475\linewidth]{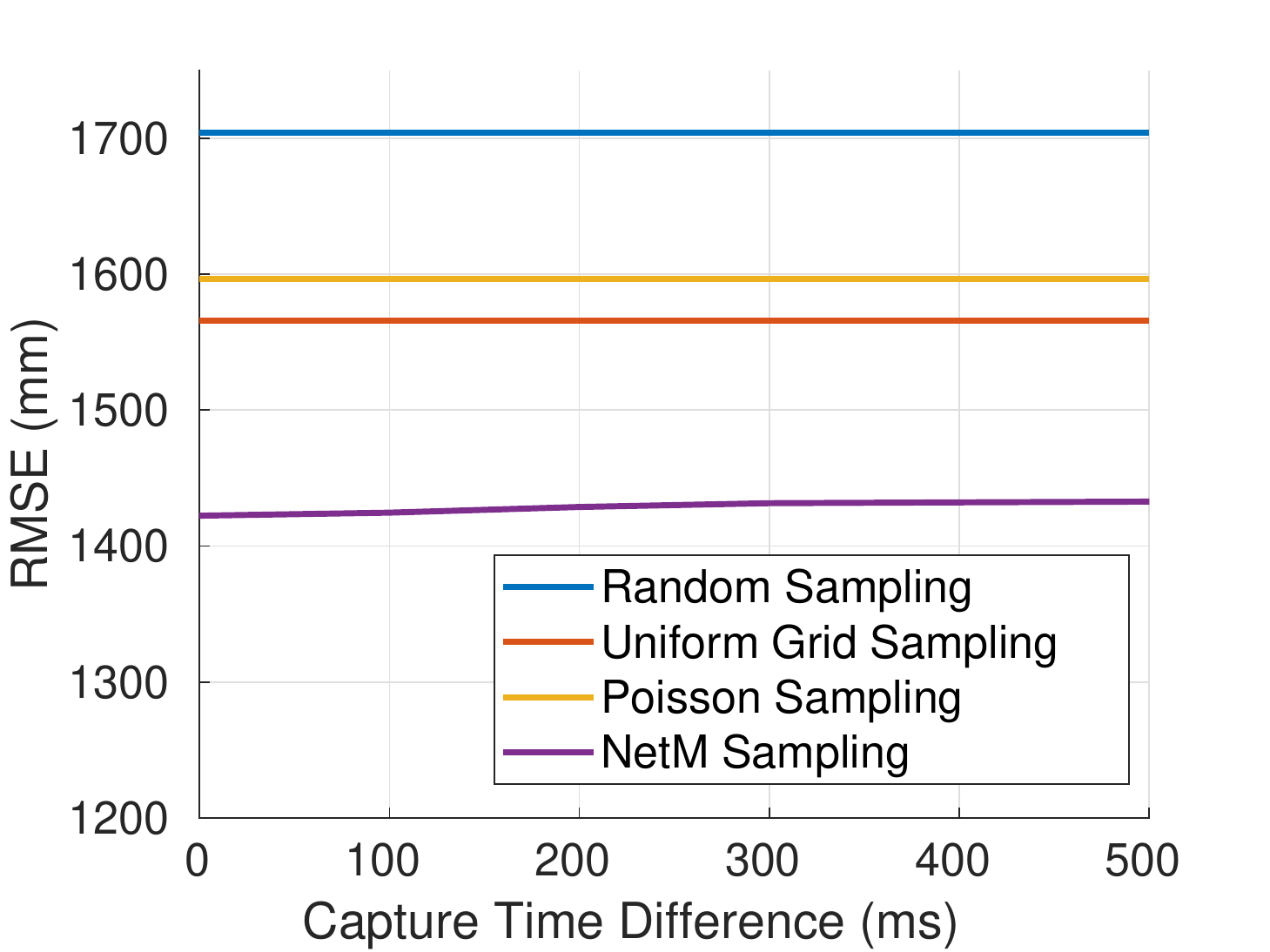}\\
\end{tabular}
\end{center}
\caption{MAE/RMSE with respect to the increasing capture time difference. Random, Uniform Grid and Poisson sampling masks do not need RGB image as input, thus they are free from the temporal registration issue.}
\label{fig:temporal_perturb}
\end{figure}

For the first set of experiments, with $0.25\%$ sampling rate on the KITTI dataset, during the depth sampling process at frame $t$, the sampling mask is computed using RGB image at frame $t-\Delta t$, to simulate the capture time difference directly. The KITTI dataset captures synchronized RGB and Depth data every $100$ms. We simulate such time delay $\Delta t$ from $0$ to $500$ms and use $NetE$ to estimate the dense depth maps. The MAE and RMSE error is shown in Figure~\ref{fig:temporal_perturb}. It can be found that $NetM$'s sampling performance is fairly stable when the capture time difference increases, up to $500$ms. One reason is that the far objects' motion is small given the temporal perturbation. Another reason is that the structure of the scene is relatively static in consecutive frames.

\begin{figure}[h]
\scriptsize
\begin{center}
\begin{tabular}{c c }
\includegraphics[width=0.475\linewidth]{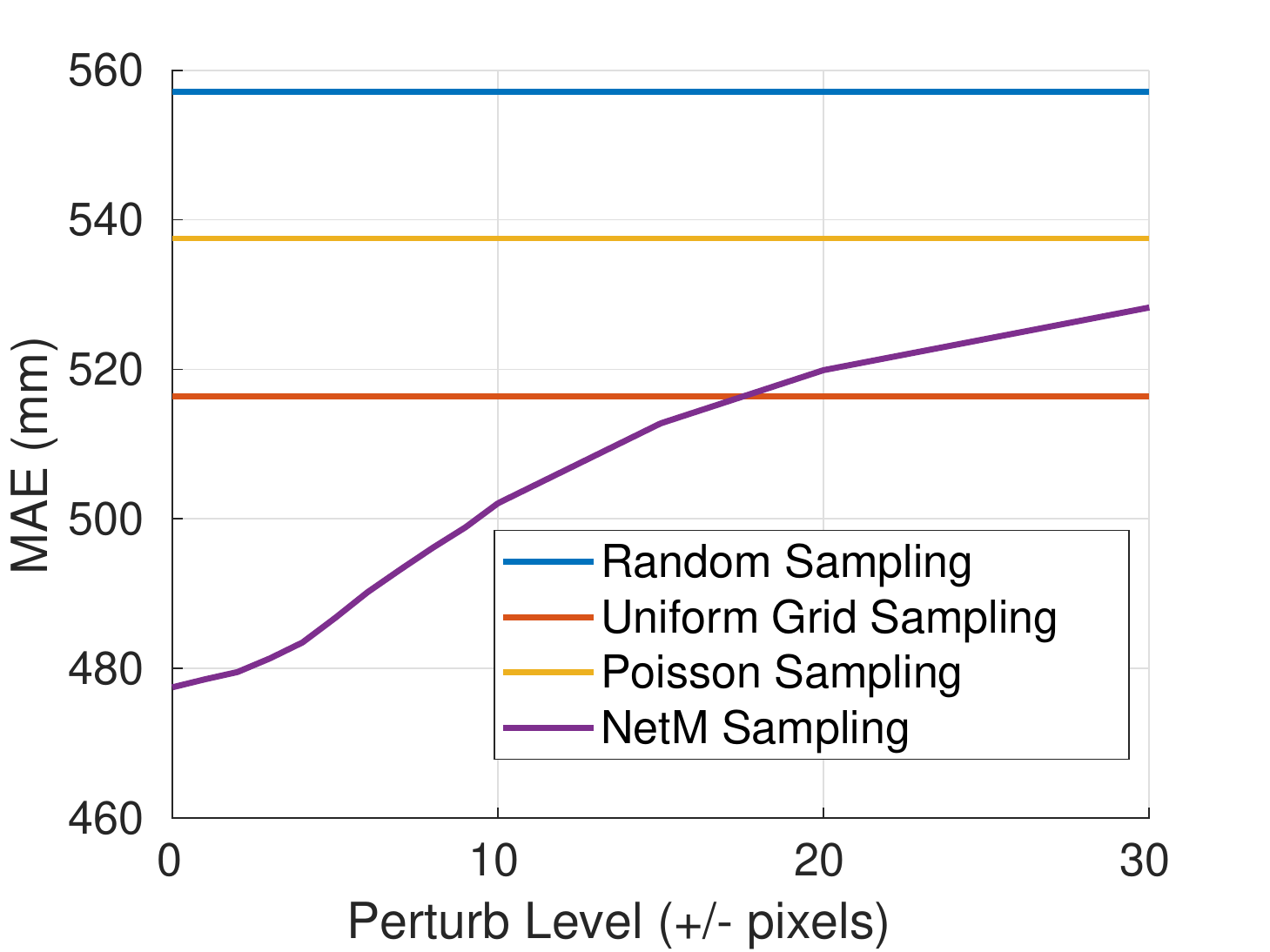} &
\includegraphics[width=0.475\linewidth]{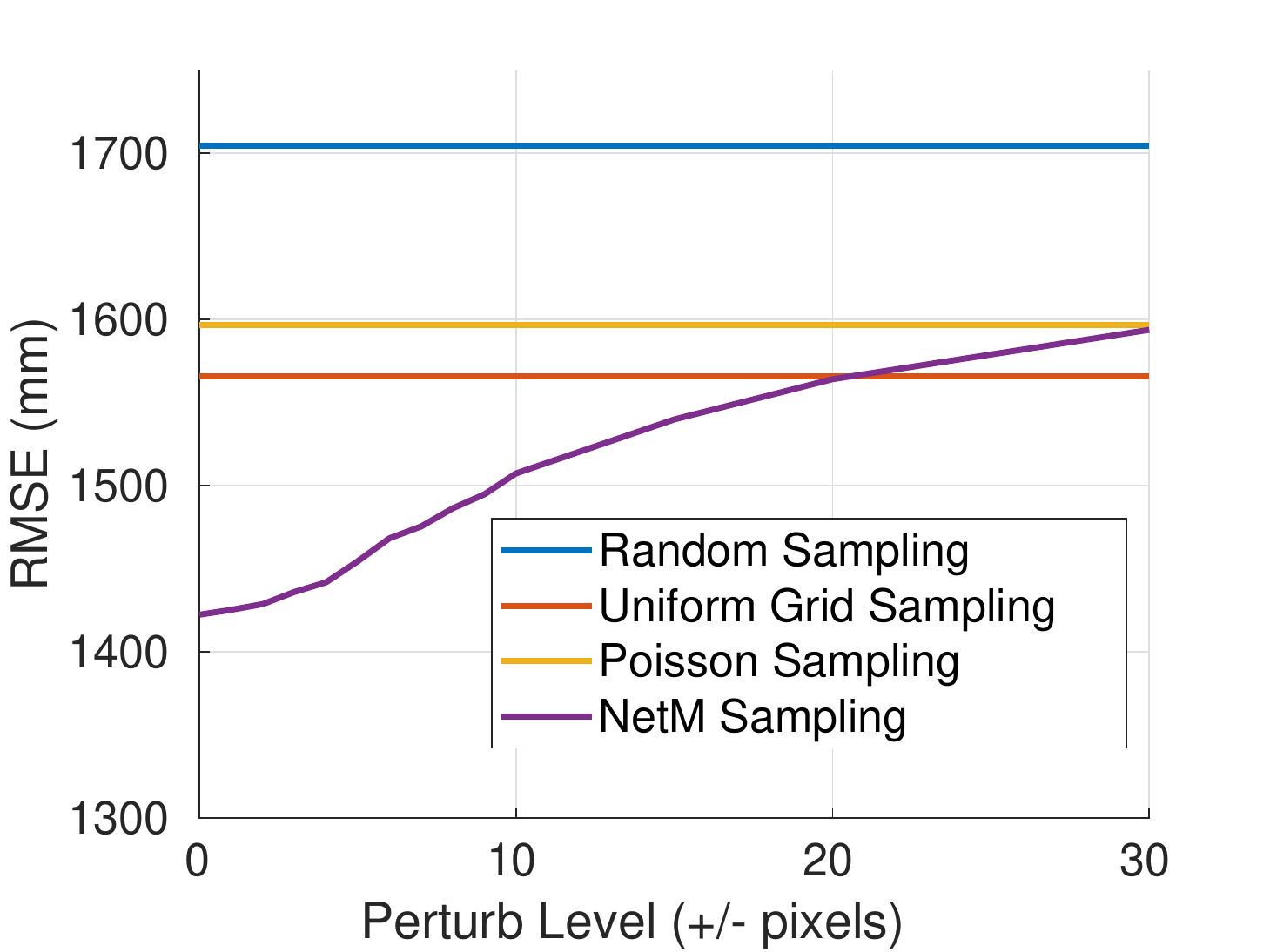}\\
\end{tabular}
\end{center}
\caption{MAE/RMSE with respect to the increasing sampling location perturbation. Random, Uniform Grid and Poisson sampling masks do not need RGB image as input, thus they are free from the location perturbation.}
\label{fig:pixel_perturb}
\end{figure}

For the second set of experiments, also with $0.25\%$ sampling rate on the KITTI dataset, additional perturbation is added on $NetM$ predicted sampling location. The perturbation is done by adding uniform distribution noise under different ranges to the sampling location. As shown in Figure~\ref{fig:pixel_perturb}, it can be found that under the MAE and RMSE metrics, even with $+/-15$ pixels perturbation ($240\times960$ full image resolution) on the sampling location, $NetM$ still outperforms Random, Uniform Grid and Poisson sampling masks (no perturbation added), under the same depth estimation method $NetE$. Such large pixel perturbation serves as a challenging test case because far objects' motion can not reach this level in practice.

According to the above 2 sets of experiments, the sampling performance advantage still holds if we take the capture time difference into consideration. The difficulty in sampling fast moving objects is one of the limitations of the proposed adaptive sampling approach. Such techniques as motion prediction can be applied to compensate the temporal registration issue. They are beyond the scope of this paper.

\section{Conclusion}
In this paper, we presented a novel adaptive depth sampling algorithm based on DL. The mask generation network $NetM$ is trained along with the depth completion network $NetE$ to predict the optimal sampling locations based on an input RGB image. Experiments demonstrate the effectiveness of the proposed $NetM$. Higher depth estimation accuracy is achieved by $NetM$ under various depth completion algorithms. We also show that best end to end performance is achieved by $NetM$ with a state-of-the-art depth completion algorithm. Such adaptive depth sampling strategy enables more efficient depth sensing and overcomes the trade-off between frame-rate, resolution, and range in an active depth sensing system (such as LiDAR and sparse dot pattern structured light sensor).

\bibliographystyle{IEEEtran}
\bibliography{egbib}

%

\begin{IEEEbiography}[{\includegraphics[width=1in,height=1.25in,clip,keepaspectratio]{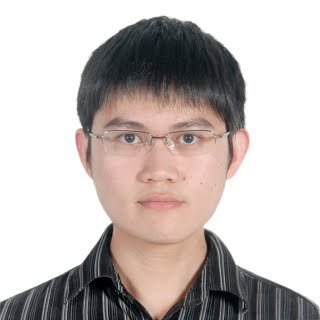}}]{Qiqin Dai}
received the B.S. degree in automation from Zhejiang University, China, in 2012. He completed his M.S. and Ph.D. degree with the Image and Video Processing Laboratory, Northwestern University, Evanston, IL, USA, in 2017. Now he is with Geomagical Labs, Mountain View, CA, USA, where he is currently working on indoor perception using machine learning. His research interests include machine learning techniques for digital image processing, computer vision and computational photography.
\end{IEEEbiography}

\begin{IEEEbiography}[{\includegraphics[width=1in,height=1.25in,clip,keepaspectratio]{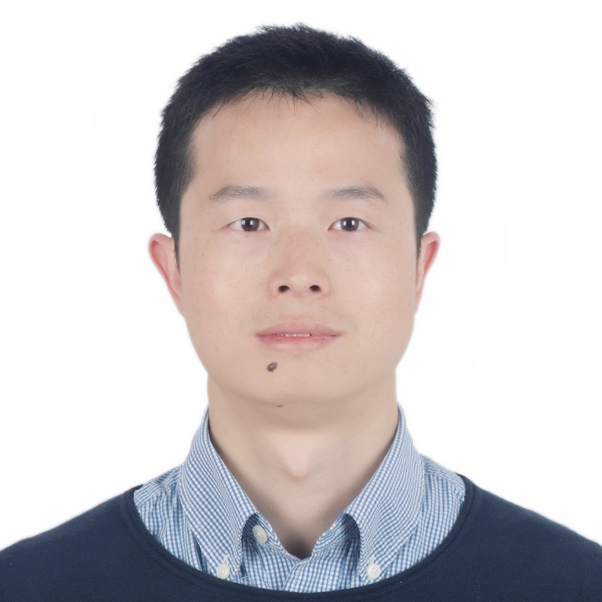}}]{Fengqiang Li}
Fengqiang Li is currently a machine learning and computer vision algorithm engineer with Apple Inc. Previously, he received his Ph.D. degree in computer science from Northwestern University. He was with Prof. Oliver Cossairt in Computational Photography Lab at Northwestern University, where he worked on computational photography and computer vision. Before that, he obtained his BS degree in optoelectronic information engineering from Huazhong University of Science and Technology and his MS degree in electrical engineering from Lehigh University.
\end{IEEEbiography}

\begin{IEEEbiography}[{\includegraphics[width=1in,height=1.25in,clip,keepaspectratio]{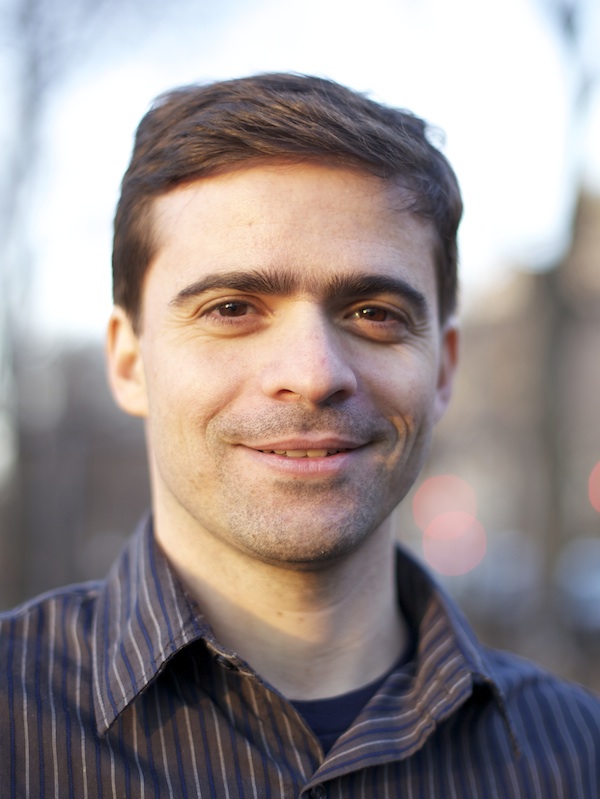}}]{Oliver Cossairt}
Oliver Cossairt is Associate Professor in the Computer Science (CS) and Electrical and Computer Engineering (ECE) departments at Northwestern University. Prof. Cossairt is director of the Computational Photography Laboratory (CPL) at Northwestern University (compphotolab.northwestern.edu), whose research consists of a diverse portfolio, ranging in topics from optics/photonics, computer graphics, computer vision, machine learning and image processing. The general goal of CPL is to develop imaging hardware and algorithms that can be applied across a broad range of physical scales, from nanometer to astronomical. This includes active projects on 3D nano-tomography (10-9 m), computational microscopy (10-6 m), cultural heritage imaging analysis of paintings (10-3 m), structured light and ToF 3D-scanning of macroscopic scenes (1 m), de-scattering through fog for remote sensing (103 m), and coded aperture imaging for astronomy (106 m). Prof. Cossairt has garnered funding from numerous corporate sponsorships (Google, Rambus, Samsung, Omron, Oculus/Facebook, Zoloz/Alibaba) and federal funding agencies (ONR, NIH, DOE, DARPA, IARPA, NSF CAREER Award).
\end{IEEEbiography}

\begin{IEEEbiography}[{\includegraphics[width=1in,height=1.25in,clip,keepaspectratio]{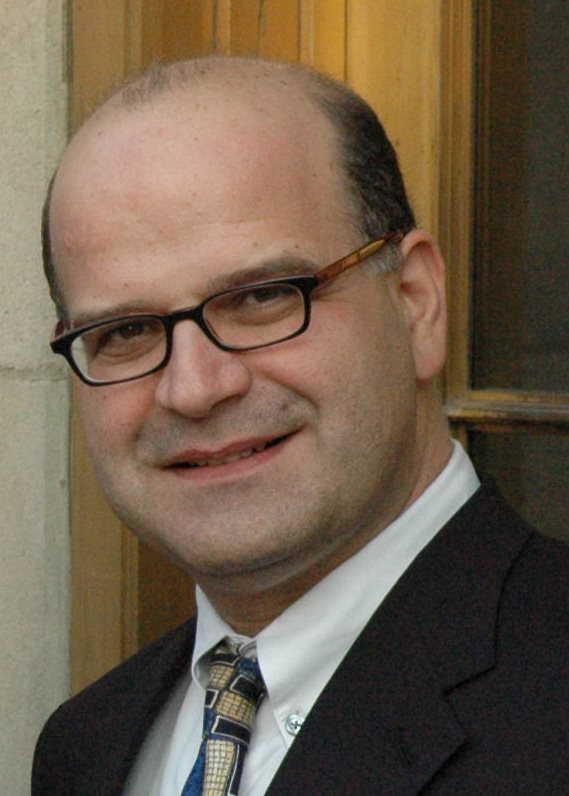}}]{Aggelos K. Katsaggelos}
received the Diploma degree in electrical and mechanical engineering from the Aristotelian University of Thessaloniki, Greece, in 1979, and the M.S. and Ph.D. degrees in Electrical Engineering from the Georgia Institute of Technology, in 1981 and 1985, respectively.
In 1985, he joined the Department of Electrical Engineering and Computer Science at Northwestern University, where he is currently a Professor holder of the Joseph Cummings chair. He was previously the holder of the Ameritech Chair of Information Technology and the AT\&T chair. He is also a member of the Academic Staff, NorthShore University Health System, an affiliated faculty at the Department of
Linguistics and he has an appointment with the Argonne National Laboratory.
He has published extensively in the areas of multimedia signal processing and communications, computational imaging, and machine learning (over 250 journal papers, 600 conference papers and 40 book chapters) and he is the holder of 30 international patents. He is the co-author of Rate-Distortion Based Video Compression (Kluwer, 1997), Super-Resolution for Images and Video (Claypool, 2007), Joint Source-Channel Video Transmission (Claypool, 2007), and Machine Learning Refined (Cambridge University Press, 2016). He has supervised 57 Ph.D. theses so far.
Among his many professional activities Prof. Katsaggelos was Editor-in-Chief of the IEEE Signal Processing Magazine (1997–2002), a BOG Member of the IEEE Signal Processing Society (1999–2001), a member of the Publication Board of the IEEE Proceedings (2003-2007), and a Member of the Award Board of the IEEE Signal Processing Society. He is a Fellow of the IEEE (1998), SPIE (2009), EURASIP (2017), and OSA (2018). He is the recipient of the IEEE Third Millennium Medal (2000), the IEEE Signal Processing Society Meritorious Service Award (2001), the IEEE Signal Processing Society Technical Achievement Award (2010), an IEEE Signal Processing
Society Best Paper Award (2001), an IEEE ICME Paper Award (2006), an IEEE ICIP Paper Award (2007), an ISPA Paper Award (2009), and a EUSIPCO paper award (2013). He was a Distinguished Lecturer of the IEEE Signal Processing Society (2007–2008).
\end{IEEEbiography}




\end{document}